\documentclass{article}

\usepackage[preprint]{Configs/NeurIPS/neurips_2026}

\usepackage[utf8]{inputenc} 
\usepackage[T1]{fontenc}    
\usepackage[hyperfootnotes=false]{hyperref}       
\usepackage{url}            
\usepackage{booktabs}       
\usepackage{amsfonts}       
\usepackage{nicefrac}       
\usepackage{microtype}      
\usepackage{graphicx}
\usepackage{svg}
\usepackage{amsmath,amssymb}
\usepackage{enumitem}
\usepackage{multirow} 
\usepackage{adjustbox}

\usepackage[dvipsnames]{xcolor}

\usepackage{Configs/mathcomEd}
\usepackage{Configs/mathstatsEd} 

\usepackage{siunitx}

\newcommand{\algo}{\textsc{Theseus}}

\usepackage{booktabs,colortbl}

\usepackage{stfloats}        

\newcommand{\term}{\mathrm{term}}
\newcommand{\rankf}{\mathrm{rank}}
\newcommand{\Lev}{\mathrm{Lev}}
\newcommand{\Edges}{\mathrm{Edges}}
\newcommand{\Relations}{\mathrm{Relations}}

\definecolor{bestcol}{RGB}{230,245,208} 
\definecolor{secondcol}{RGB}{255,247,188} 
\definecolor{Burgundy}{RGB}{144,0,32}
\newcommand{\best}[1]{\textbf{#1}}

\newcommand{\eduin}{ \color{ForestGreen}}

\title{
    Theseus in the Graph: \\ Towards Traceable Multi-Hop Graph Navigation
}

\author{
    Eduin E.~Hernandez$^{1}$\thanks{Corresponding author: \texttt{eduin.ee08@nycu.edu.tw}.}\And
    Luis F.~Garcia$^{2}$ \And
    Nurassyl Askar$^{2}$ \And
    Sergio A.~Diaz$^{1}$ \And 
    Stefano Rini$^{1}$ \\
    $^{1}$ National Yang Ming Chiao Tung University, Hsinchu, Taiwan \\
    $^{2}$ Independent Researcher \\
}

\begin{document}

\maketitle

\begin{abstract}
Multi-Hop Knowledge Graph Question Answering (KGQA) tasks require models to assemble relational evidence along paths in a KG to answer natural-language questions. 
However, existing KGQA systems typically focus on predicting the final answer without explicitly modeling or validating the intermediate reasoning steps, obscuring whether the correct answers arise from faithful multi-hop reasoning.
To address this limitation, we re-frame multi-hop KGQA as a \emph{question-conditioned graph navigation} problem.
We refer to this formulation as \textsc{Theseus} -- Traceable Hop-wise Evidence SEarch in a Unified Semantics. 
%
%
In this setting, an agent receives a KG, a question, and a topic entity, and traverses a sequence of relations towards the answer, making the reasoning path explicit. 
To systematically study this formulation, we provide three key contributions. 
(i) We augment the existing \textsc{KINSHIP} and \textsc{MQuAKE} resources into navigation-ready KGQA datasets with annotated evidence paths and paraphrased questions.  
(ii) We design evaluation protocols to measure path fidelity, robustness to linguistic variation, and performance across multi-hop and multi-answer questions. 
(iii) We adapt established path-based KG completion agents—\textsc{MINERVA}, \textsc{MultiHopKG}, and \textsc{SQUIRE}—to operate on full question embeddings rather than symbolic single-relation queries, enabling their trajectories to be guided by natural-language semantics.
%
Together, these contributions advance KGQA research toward systems where traceability is fundamental: answers are accompanied by explicit reasoning paths whose agreement with reference evidence can be systematically evaluated.
\end{abstract}

\section{Introduction}
\label{section:introduction}

Humans phrase questions with remarkable efficiency: by drawing on shared context, logic, and implicit knowledge, a single query can encode a rich hierarchy of meaning.
Natural‑language questions thus lend themselves to being decomposed into simpler semantic units—facts, relations, and reasoning steps—which aligns neatly with the structure of knowledge graphs.
Motivated by this observation, we model question answering as question‑conditioned navigation: given a graph, a topic entity, and a question, the model follows a chain of relations, guided by the question’s semantics, to reach the answer.
We refer to this formulation as the \textsc{Theseus} setting, viewing the KG as a labyrinth: entities are nodes, relations are pathways, the topic entity is the entrance, the answer is the exit, and the question defines the route.
A notable advantage of this framing is traceability: each reasoning step corresponds to a discrete relation, so the resulting path itself becomes part of the explanation.
\textsc{Theseus} could in principle provide actionable insight rather than merely delivering an answer; for example, when diagnosing a malfunctioning device, a traceable KGQA system can guide a user through each check and adjustment, providing both the solution and the reasoning behind it.
In this paper, we introduce datasets, evaluation protocols, and models for \textsc{Theseus}, and argue that explicit, inspectable reasoning trajectories are an important component of practical multi-hop KGQA.

\paragraph{Problem Statement:}
We study multi-hop KGQA as a question-conditioned graph navigation problem and refer to this formulation as \textsc{Theseus} -- {T}raceable {H}op-wise {E}vidence {SE}arch in a {U}nified {S}emantics.
The ``unified semantics'' refers to the shared representation in which the natural-language question \(q\) and KG elements are aligned, enabling the semantics of \(q\) to guide the choice of relations and entities at each step.
Let $\mathcal{G} = (\mathcal{E},\mathcal{R},\mathcal{T})$ denote a KG, where $\mathcal{E}$ is a set of entities, $\mathcal{R}$ is a set of relations, and $\mathcal{T} \subseteq \mathcal{E} \times \mathcal{R} \times \mathcal{E}$ is the set of triples.
Given a natural‑language question $q$, a starting/topic entity $e_s \in \mathcal{E}$, and $\mathcal{G}$, the task is to produce a sequence
\[
(e_0 = e_s, r_1, e_1, r_2, e_2, \dots, r_n, e_n = e^\star)
\]
such that $(e_{i-1}, r_i, e_i) \in \mathcal{T}$ for all $i \in \{1,\dots,n\}$ and $e^\star$ is an entity that satisfies $q$.
%
%
The underlying evidence-path length $n$ is not provided to the agent in advance. 
Instead, the agent operates under a fixed reasoning horizon $N \ge n$, allowing questions with different underlying hop lengths to be evaluated under a common traversal budget.
%
%
This formulation yields a path-structured answer: the reasoning path itself is part of the output, providing an explicit and traceable record of how the agent reached the answer.
In this sense, \textsc{Theseus} bridges two lines of research that are often treated separately: (i) knowledge‑graph completion (KGC), which aims to predict whether a triple exists in a KG, and (ii) knowledge‑graph question answering (KGQA), which conditions inference on natural‑language questions. 
In the next section, we review these two lines of work.

\subsection{Related Work}
\label{subsec:related_work}

\begin{figure*}[t]
    \centering
    \makebox[\textwidth][c]{%
        \adjustbox{
            max width=0.8\textwidth,
            trim=1cm 0cm 13cm 1cm,
            clip
        }{%
            \includegraphics{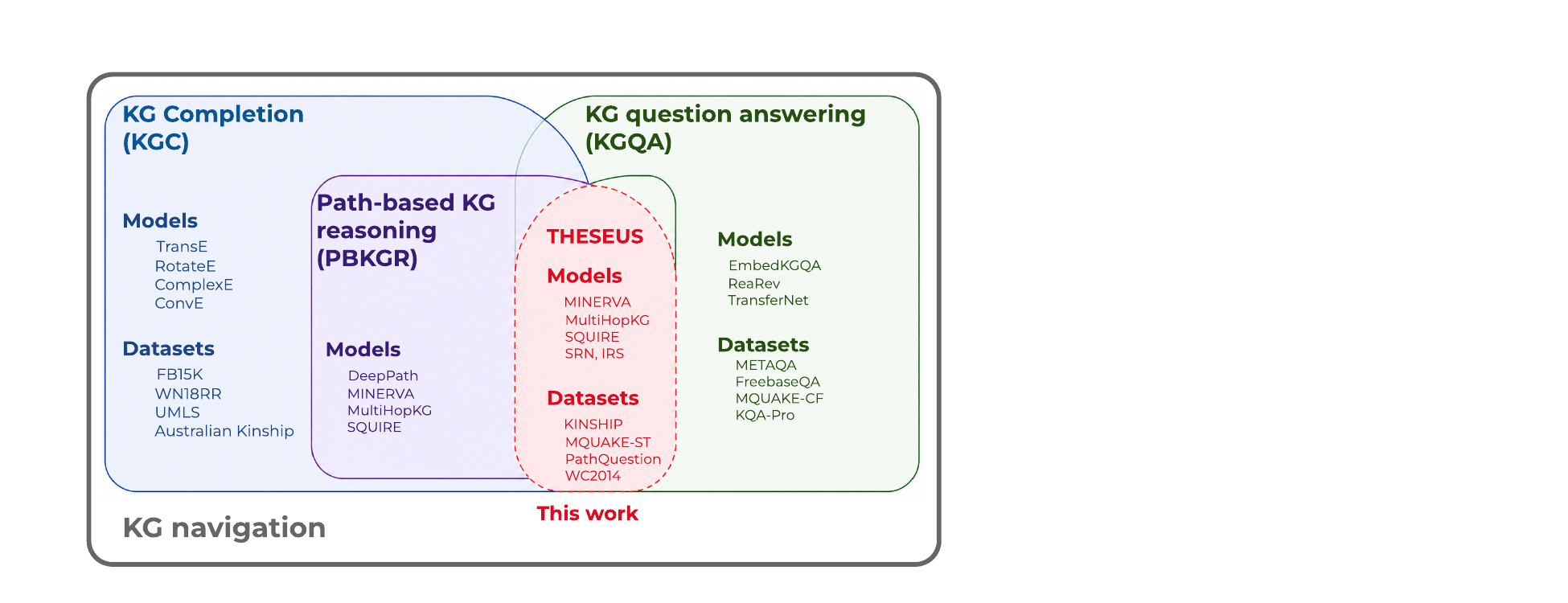}%
        }%
    }
    \caption{
    Taxonomy of KG Navigation settings.
    \algo \ lies at the intersection of KGQA and path-based KG reasoning, extending explicit KG navigation to natural-language questions and path-level evaluations.
    }
    \label{fig:diagram}
\end{figure*}

We begin by situating our work within a taxonomy of graph-navigation research, which helps delineate our contributions and is summarized in Fig.~\ref{fig:diagram}.

Broadly, graph navigation can be divided into two areas: \emph{knowledge‑graph completion} (KGC) and \emph{knowledge‑graph question answering} (KGQA). 
KGC methods were introduced first and aim to predict whether a particular triple \((h,r,t)\) belongs in a KG. 
%
%
A subset of KGC methods perform \emph{path‑based KG reasoning} (PBKGR): here the agent must construct a path from a head entity to a tail entity consistent with the given relation. 
Our work lies at the intersection of KGC and KGQA: like PBKGR, we predict explicit reasoning paths, but we condition these paths on natural‑language questions as in KGQA.

After reviewing embedding‑based and path‑based methods in both KGC and KGQA, we discuss existing graph‑navigation datasets and how they have been augmented to support KGC, KGQA, or PBKGR. 
We conclude by highlighting the supervision and reproducibility properties needed for path‑level evaluation in KGQA.

\smallskip
\noindent
\textbf{KG Completion (KGC):}

\noindent
$\bullet$ \emph{Embedding-based Entity Ranking.}
A large body of work studies KGC via embedding-based scoring functions for link prediction and entity ranking~\cite{bordes2013translating,sun2019rotate,yang2015embedding,trouillon2016complex,dettmers2018convolutional}.
Given an atomic query pattern such as $(\text{head},\text{relation},?)$, these models score candidate tail entities and rank them according to their plausibility, thereby recovering missing facts in an incomplete KG.
%
%
Despite the very limited input format, embedding-based KGC models can provide useful entity--relation representations or scoring functions that are reused in downstream KGQA and path-based reasoning models~\cite{lin2018multi,qiu2020stepwise,saxena2020improving}.

\noindent
$\bullet$ \emph{Path-based KG reasoning (PBKGR):}
Path-based approaches to KGC and reasoning formulate inference as a sequential decision-making problem over the local neighborhoods of a KG.
\textsc{DeepPath}~\cite{xiong2017deeppath} learns to discover predictive multi-hop paths connecting a given head--tail entity pair for relation reasoning, whereas \textsc{MINERVA}~\cite{das2018go} addresses symbolic query answering of the form $(\text{head},\text{relation},?)$ by training a reinforcement learning agent to walk from a head entity toward an unknown tail entity.
Building on this paradigm, \textsc{MultiHopKG}~\cite{lin2018multi} augments reinforcement-learning-based KG reasoning with embedding-based reward shaping, using a pretrained KG embedding model to provide softer rewards for unobserved facts.
%
\textsc{SQUIRE}~\cite{bai2022squire} departs from RL-based walking by formulating multi-hop KG reasoning as a sequence-to-sequence problem that generates complete evidential path sequences for symbolic triple queries.
Beyond answer accuracy, \cite{lv2021multi} evaluate the interpretability of answer-reaching paths in multi-hop link prediction using Path Recall and human-annotated rule-level interpretability scores.
Their evaluation is complementary to our path-fidelity diagnostics: they assess whether discovered reasoning paths are human-interpretable, whereas our metrics measure structural agreement with annotated reference evidence paths; Appendix~\ref{app:path_fidelity_metrics} provides a detailed comparison.
Although these methods expose explicit reasoning paths, they are primarily designed for KGC, where the input is a structured query rather than a natural-language multi-hop question.
Consequently, they primarily focus on answer metrics.

\smallskip
\noindent
\textbf{KG question answering (KGQA):}

\noindent
$\bullet$ \emph{Direct Answer Prediction with Latent Reasoning.}
A broad class of Multi-Hop KGQA methods targets answer prediction by mapping a natural-language question, often together with linked topic entities and/or a question-specific graph context, to scores over candidate answer entities while leaving the underlying evidence path implicit.
One strategy is to reuse KG embedding models for question answering:
\textsc{EmbedKGQA}~\cite{saxena2020improving} encodes the question and topic entity in a pretrained KG embedding space and directly scores candidate answer entities from the entity set of the input KG.
By avoiding answer selection from a pre-specified local neighborhood, this formulation improves robustness when supporting KG links are sparse or missing.
Another line constructs a question-specific evidence graph and performs differentiable reasoning over it.
\textsc{ReaRev}~\cite{mavromatis2022rearev} assumes topic entities and a pre-extracted question-specific subgraph as input, and performs GNN-based reasoning guided by adaptively updated question instructions to score candidate answer nodes within that subgraph.
Transfer-oriented approaches such as \textsc{TransferNet}~\cite{shi2021transfernet} perform differentiable multi-hop reasoning by propagating entity scores across question-activated relations, supporting both labeled KG relations and textual relations in a unified relation graph.
Although these models may expose intermediate node distributions or entity scores, their prediction target remains the answer entity rather than an explicit evidence path.
%

\noindent
$\bullet$
\emph{Path-based KG reasoning (PBKGR):} 
Closer to our setting, path-based KGQA methods treat question answering as a step-wise reasoning process over KG structure rather than as purely latent answer prediction.
The Interpretable Reasoning Network (IRN) performs hop-by-hop reasoning by predicting a relation and an entity at each step, producing traceable intermediate predictions for multi-relation questions~\cite{zhou2018interpretable}.
The Stepwise Reasoning Network (SRN) formulates multi-relation KGQA as a weakly supervised sequential decision-making problem, where a reinforcement-learning agent searches over KG paths from the topic entity under the guidance of the input question~\cite{qiu2020stepwise}.
These models demonstrate that natural-language questions can guide explicit reasoning over KG structure, making them closely related to our \emph{Question-Conditioned Graph Navigation} setting.
Their analyses also show the value of inspecting intermediate reasoning steps; however, such evaluations are largely model-specific, making it difficult to compare answer correctness and reasoning-path fidelity under a common protocol.
This motivates a setting in which both the reached answer and the executed reasoning trajectory can be evaluated systematically.

\subsection{Relevant Datasets}
\label{subsubsec:related_datasets}


\noindent
\textbf{Multi-hop and compositional KGQA:}
Several KGQA benchmarks support compositional reasoning, but differ in the form of supervision they provide and in how directly they support path-level graph-navigation evaluation.
\textsc{MetaQA}~\cite{zhang2017variational} provides large-scale 1--3-hop questions over a movie-domain KG, but does not provide per-question evidence paths, limiting direct evaluation of whether a predicted trajectory follows the intended reasoning chain.
\textsc{KQA-Pro}~\cite{cao2022kqa} provides executable KoPL/SPARQL programs covering multi-hop reasoning and other logical operations, making it well suited to semantic parsing and program-execution evaluation.
Similarly, \textsc{WebQSP}~\cite{yih2016value} pairs natural-language questions with executable SPARQL queries over Freebase, while \textsc{ComplexWebQuestions}~\cite{talmor2018web} extends this setting to more compositional queries involving composition, conjunction, comparison,
and superlatives.
These resources provide rich query-level supervision, but their reasoning targets are executable programs or logical query structures rather than single ordered entity--relation trajectories.
They are therefore complementary to path-based navigation benchmarks, which instead require evaluating the sequence of graph transitions executed from a topic entity to an answer.
\textsc{FreebaseQA}~\cite{jiang2019freebaseqa} provides trivia-style questions aligned primarily to individual Freebase triples and therefore serves a different, largely single-hop KGQA setting.
Because \textsc{WebQSP}, \textsc{ComplexWebQuestions}, and \textsc{FreebaseQA} are grounded in Freebase~\cite{bollacker2008freebase}, reproducing experiments against the original KG can additionally depend on the availability of compatible Freebase snapshots.

\noindent
\textbf{PBKGR Datasets:}
\textsc{PathQuestion} and \textsc{PathQuestion-Large}~\cite{zhou2018interpretable} and \textsc{WC2014}~\cite{zhang2016gaussian} provide questions grounded in explicit reasoning paths and are therefore closely aligned with our setting.
Their standard evaluation protocols, however, use compact task-specific
navigation KGs that differ from our shared-graph setting in several structural
and traversal characteristics; Appendix~\ref{app:prior_dataset} quantifies
these differences using the same diagnostics as our datasets.
\textsc{MQuAKE-CF}~\cite{zhong2023mquake} provides annotated 2--4-hop questions and evidence paths for studying knowledge-editing effects, but does not release a fixed navigation KG.
As detailed in Appendix~\ref{app:mquake_st_extra}, constructing a KG only from its annotated paths produces a highly fragmented graph, whereas augmenting it with external triples introduces dependence on the augmentation procedure and KG snapshot.
These limitations motivate navigation-ready KGQA resources that combine explicit topic entities and evidence paths with a materialized shared KG snapshot and interpretable entity/relation mappings for reproducible traceability analysis.

\subsection{Contributions}
\label{subsec:contributions}

We formalize \emph{question-conditioned graph navigation} as the \textsc{Theseus} setting: \textbf{T}raceable \textbf{H}op-wise \textbf{E}vidence \textbf{Se}arch in a \textbf{U}nified \textbf{S}emantics.
Given a KG, a question $q$, and a start entity $e_s$, an agent must navigate toward the answer without being told the required hop length.
If the underlying evidence path has length $n$, the agent is evaluated under a reasoning horizon $N \ge n$, separating the intrinsic complexity of the question from the exploration budget available to the agent.
Our contributions are:
\begin{itemize}[leftmargin=*,noitemsep]
    \item \textsf{[Dataset]} -- \textbf{Navigation-ready datasets with gold paths.}
    We construct two KGQA datasets, \textsc{Kinship}~\cite{kinship_55} and \textsc{MQuAKE-ST} (our static-KG variant of \textsc{MQuAKE-CF}~\cite{zhong2023mquake}), providing explicit start entities, frozen navigation graphs, annotated evidence paths, and controlled linguistic variations for evaluating robustness to paraphrasing.
    
    \item \textsf{[Metrics]} -- \textbf{Systematic evaluation and structural calibration of reasoning behavior.}
    We propose evaluation protocols that jointly measure \emph{answer accuracy} and \emph{path fidelity}, and complement these metrics with question-agnostic structural calibration references based on unbiased random walks and answer-oracle shortest paths.
    Together, these diagnostics help distinguish reference-path agreement attributable to question conditioning from agreement or answer reachability that can arise from graph structure alone.
    
    \item \textsf{[Agent]} -- \textbf{Question-conditioned path-based navigation agents.}
    We adapt representative navigators---\textsc{MINERVA}~\cite{das2018go}, \textsc{MultiHopKG}~\cite{lin2018multi}, and \textsc{SQUIRE}~\cite{bai2022squire}---from symbolic link-prediction queries to question-conditioned graph navigation, enabling step-wise traversal conditioned on the semantic alignment between natural-language questions and KG embeddings.
\end{itemize}
%

Code, datasets, pretrained checkpoints, and evaluation resources are available through the \textsc{Theseus} project repository.\footnote{\url{https://github.com/HalcyonSolutions/THESEUS}}

\section{Preliminaries}
\label{section:preliminaries}

\noindent \textbf{Multi-Hop KG Navigation:}
Given a start entity $e_s \in \mathcal{E}$ and a natural-language question $q$, multi-hop navigation requires composing a sequence of relational edges $r_1, r_2, \dots, r_n$ to reach an answer entity $e^\star$.
Let $e_0 = e_s$ and $e_n = e^\star$; then a valid reasoning path $\Psf$ is defined as 
\begin{equation}
    \Psf=\big( (e_0, r_1, e_1),\; (e_1, r_2, e_2), \; \dots,\; (e_{n-1}, r_n, e_n) \big).
    \label{eq:path}
\end{equation}
The corresponding relation sequence is defined as
\begin{equation}
    \Rsf(\Psf) = (r_1,r_2,\dots,r_n),
    \label{eq:relation_sequence}
\end{equation}
which captures the ordered relational composition of the path while ignoring the intermediate entities.
The integer $n$ in \eqref{eq:relation_sequence} specifies the number of hops in the reasoning process.
Multi-hop navigation thus amounts to selecting and executing such a path in response to a given question.

\smallskip
\noindent \textbf{KG Embedding:}
We assume entities and relations in $\mathcal{G}$ may be represented in 
an embedding space.
For real-valued KGE models, let $\mathbf{E}\in\mathbb{R}^{|\mathcal{E}|\times d_{\mathrm{KG}}}$ and $\mathbf{R}\in\mathbb{R}^{|\mathcal{R}|\times d_{\mathrm{KG}}}$ denote entity and relation embedding matrices obtained by stacking the entity and relation embeddings $\{\ev_i\}_{i \in \Ecal}$ and $\{\rv_j\}_{j \in \Rcal}$, respectively.
For complex-valued KGE models,
the corresponding embeddings instead lie in $\mathbb{C}^{d_{\mathrm{KG}}}$.
For a triple $(u,r,v)$, a KG scoring function
\begin{equation}
    s(u,r,v)=g(\mathbf{e}_u,\mathbf{r},\mathbf{e}_v),
\end{equation}
assigns structural plausibility to the triple.
These scores can be used as structural priors for ranking outgoing actions and are typically learned via link prediction.
Common KG scoring functions, including \emph{TransE}, \emph{DistMult}, \emph{ComplEx}, \emph{RotatE}, and \emph{ConvE}, are summarized in Appendix~\ref{app:kg_embedding_scoring}.

\subsection{Navigation Agent}
\label{sec:Navigation Agent}
We model navigation as discrete decisions over the neighborhood of the current node. 
%
%
At step $t$ the state is $s_t=(e_t, \mathbf{m}_t , c)$, where $e_t\in\mathcal{E}$ is the current entity, $\mathbf{m}_t$ is a recurrent memory summarizing the trajectory history, and $c$ is the \emph{conditioning context}.
Let the labeled edges from $e_t$ be
\begin{equation}
  \mathcal{N}(e_t)\;=\;\{\, (r,e') \;:\; (e_t,r,e')\in\mathcal{T}\,\}.
\end{equation}
A scoring function $f_\theta$ produces the logits, defining the policy and action space as
\eas{
\pi_\theta(a_t\mid s_t)& = \sigma\!\big(f_\theta(e_t,a_t,\mathbf{m}_t;c)\big) \\
a_t& =(r_{t+1},e_{t+1})\in\mathcal{N}(e_t),
}{\label{eq:agent}}
where $\sigma$ indicates the softmax operation.
Episodes proceed for a fixed hop budget $N$ with $N \ge n$ (where $n$ is the annotated path length); revisits (cycles) are allowed.
For a trajectory $\tau=(a_0,\ldots,a_{N-1})$ starting at $e_s$, the log-probability is
\begin{equation}
  \log p_\theta(\tau\mid e_s,c)\;=\;\sum_{t=0}^{N-1}\log \pi_\theta(a_t\mid s_t).
\end{equation}
We denote the \emph{answer set} for an instance by
\begin{equation}
  \mathcal{A}(q)\;\subseteq\;\mathcal{E},
\end{equation}
i.e., the set of entities that are considered correct answers to the query/question $q$.

Following the taxonomy in Sec. \ref{subsec:related_work}, we specialize the agent in Eq.\eqref{eq:agent} in the two following settings.

\smallskip
\noindent
\textbf{KGC agent:}
%
In the conventional atomic setting, the query $q$ is the relation itself:
\[
q \;:=\; r,\qquad c\;=\;r,\qquad e_s=u,
\]
and the answer set is
\begin{equation}
  \mathcal{A}(q)\;=\;\{\,v\in\mathcal{E}\;:\;(u,r,v)\in\mathcal{T}\,\}.
\end{equation}
The policy is conditioned only on $r$ and navigates from $u$; to avoid trivial leakage, the target triple $(u,r,v)$ is masked from the immediate choices when constructing the instance.

\smallskip
\noindent
\textbf{KGQA agent:}
%
To represent natural-language questions, we encode $q =  (w_1,\ldots, w_m)$ with a pre-trained \emph{language model} (e.g., BERT \cite{devlin2019bert}) to obtain token embeddings $\mathbf{z}_i\in\mathbb{R}^{d_{\mathrm{LM}}}$.
We then pool these embeddings to obtain a fixed-size vector and \emph{project} it into the KG embedding space $\mathbb{R}^{d_{\mathrm{KG}}}$ (the space of entity/relation embeddings):
\[
  \mathbf{z}_q=\mathrm{Pool}(\mathbf{z}_1,\ldots,\mathbf{z}_m)\in\mathbb{R}^{d_{\mathrm{LM}}},
\]
\[
  \tilde{\mathbf{z}}_q = \phi(\mathbf{z}_q)\in\mathbb{R}^{d_{\mathrm{KG}}},
\]
where $\phi:\mathbb{R}^{d_{\mathrm{LM}}}\!\to\!\mathbb{R}^{d_{\mathrm{KG}}}$ is a \emph{trainable projection function} that maps the question into the common latent space used by KG embeddings ($\mathbf{e},\mathbf{r}\in\mathbb{R}^{d_{\mathrm{KG}}}$).
We then set the conditioning context and start entity as
\[
  c=\tilde{\mathbf{z}}_q,
  \qquad
  e_s \text{ is specified in the instance},
\]
\begin{equation}
  \mathcal{A}(q)\subseteq\mathcal{E} \text{ (answer set provided by the dataset)}.
\end{equation}
Here $c$ encodes the full question semantics \emph{in the shared KG–language-model space} used to guide path selection.
This formulation provides the common framework used to adapt navigation-based KGC agents to KGQA.

Model-specific adaptation details for \textsc{MINERVA}, \textsc{MultiHopKG}, and \textsc{SQUIRE} are provided in Appendix~\ref{app:model_adaptations}.

\section{Problem Formulation}
\label{sec:problem_formulation}

Echoing the discussion in Sec. \ref{subsec:contributions}, we define the \algo \ setting by the definition of (i)  dataset, (ii) metric, and (iii) agent.

\smallskip
\noindent
$\bullet$ \textsf{[Dataset]:}
To support \textsc{Theseus}, a dataset must provide question-answer pairs grounded in a fixed, curated KG shared across all questions.
Each instance should include an explicit topic entity, one or more answer entities, and reference-path supervision sufficient to identify one or more valid evidence paths connecting the topic entity to the valid answer entities.
The KG should also provide human-readable entity and relation labels, ensuring that the surface forms used in the questions can be linked consistently to the corresponding topic, answer, and relation identifiers in the KG.
Finally, the dataset should include controlled linguistic variations of the questions to evaluate robustness under rephrasing.

\smallskip
\noindent
$\bullet$ \textsf{[Metrics]:}
For each instance $(q,e_s)$, the evaluated method generates $B_{\mathrm{roll}}$ scored candidate trajectories under hop budget $N$.
For answer evaluation, trajectories terminating at the same entity are deduplicated by retaining the highest score, and the resulting terminal entities are ranked.
We report Hits@$K$ and MRR over the resulting ranked candidate entities
with respect to the valid answer set $\mathcal{A}(q)$.
The complete candidate-generation and answer-ranking procedure is provided in Appendix~\ref{app:rollout_ranking}.

\smallskip
\noindent\textbf{Path-fidelity metrics:}
Beyond whether an agent reaches a valid answer, \textsc{Theseus} evaluates whether its preferred trajectory agrees with the reference reasoning structure.
Let $\widehat{\Psf}(q)$ denote the path induced by the highest-scoring trajectory and $\Psf^\star(q)$ a reference evidence path.
We measure ordered path and relation agreement using
\begin{equation}
\begin{aligned}
    \mathrm{PED}(q)
    &=
    \Lev\!\left(
        \widehat{\Psf}(q),\Psf^\star(q)
    \right),\\
    \mathrm{RED}(q)
    &=
    \Lev\!\left(
        \Rsf(\widehat{\Psf}(q)),
        \Rsf(\Psf^\star(q))
    \right),
\end{aligned}
\label{eq:trace_edit}
\end{equation}
where $\Lev(\cdot,\cdot)$ is the Levenshtein distance and $\Rsf(\cdot)$ extracts the ordered relation sequence.
We complement these ordered metrics with the order-invariant overlap scores
\begin{equation}
\begin{aligned}
    \mathrm{F1}_{\mathrm{SG}}(q)
    &=
    \mathrm{F1}\!\left(
        \Edges(\widehat{\Psf}(q)),
        \Edges(\Psf^\star(q))
    \right),\\
    \mathrm{F1}_{\mathrm{Rel}}(q)
    &=
    \mathrm{F1}\!\left(
        \Relations(\widehat{\Psf}(q)),
        \Relations(\Psf^\star(q))
    \right),
\end{aligned}
\label{eq:trace_overlap}
\end{equation}
where $\Edges(\cdot)$ and $\Relations(\cdot)$ denote the traversed edge and relation sets, respectively, and $\mathrm{F1}(\cdot,\cdot)$ denotes their set-overlap F1 score.
Thus, $\mathrm{PED}$ and $\mathrm{F1}_{\mathrm{SG}}$ are entity-aware, comparing complete edges, whereas $\mathrm{RED}$ and $\mathrm{F1}_{\mathrm{Rel}}$ compare only relation labels; within each pair, the edit-distance metric preserves order while the F1 metric measures order-invariant overlap.
Full definitions, including the overlap computation, dataset-level aggregation, and multi-answer extension, are provided in Appendix~\ref{app:path_fidelity_metrics}.

\smallskip
\noindent
$\bullet$ \textsf{[Agent]:} For \textsc{Theseus}, we consider a KGQA agent as defined in Sec.  \ref{sec:Navigation Agent}.

\section{Dataset}
\label{section:datasets}

\begin{table}[b]
    \caption{Structural KG statistics.}
    \label{tab:kg_stats}
      \centering
      \begin{tabular}{l | c c c}
        \toprule
        \textbf{Dataset} & \textbf{\textsc{Kinship}} & \textbf{\textsc{MetaQA}} & \textbf{\textsc{MQuAKE-ST}} \\
        \midrule
        Entities $|\mathcal{E}|$ & 24 & 43{,}234 & 38{,}516 \\
        Relations $|\mathcal{R}|$ & 12 & 9 & 665 \\
        Triples  $|\mathcal{T}|$ & 112 & 134{,}741 & 724{,}141 \\
        Avg. Out-Degree & 4.67 & 3.12 & 18.80 \\
        WCCs & 2 & 30 & 1 \\
        Graph Density & \num[round-mode=places, round-precision=2, scientific-notation=true]{2.029E-01} & \num[round-mode=places, round-precision=2, scientific-notation=true]{1.334E-04} & \num[round-mode=places, round-precision=2, scientific-notation=true]{8.663E-04} \\
        Clustering Coefficient  & \num[round-mode=places, round-precision=2, scientific-notation=true]{7.651E-01} & \num[round-mode=places, round-precision=2, scientific-notation=true]{3.908E-04} & \num[round-mode=places, round-precision=2, scientific-notation=true]{1.501E-01} \\
        \bottomrule
      \end{tabular}
\end{table}

\begin{table}
    \caption{Number of questions per hop length and KGQA traversal difficulty. Question counts are computed over the full QA set, while traversal difficulty is evaluated on the test split for the corresponding $n$-hop setting.}
    \label{tab:qa_stats}
    \centering
    \small
    \resizebox{\textwidth}{!}{
    \begin{tabular}{l c | r r r r r | c c c}
        \toprule
        \multirow{2}{*}{\textbf{Dataset}}
        & \multirow{2}{*}{\textbf{Answer Type}}
        & \multicolumn{5}{c|}{\textbf{Question Counts by Hop Length}}
        & \multicolumn{3}{c}{\textbf{Traversal Difficulty}} \\
        \cmidrule(lr){3-7}
        \cmidrule(lr){8-10}
        &
        & \textbf{1-Hop}
        & \textbf{2-Hop}
        & \textbf{3-Hop}
        & \textbf{4-Hop}
        & \textbf{Total}
        & $\textbf{RW-Ans}_p$
        & $\textbf{RW-Ans}_{\mathrm{MC}}$
        & \textbf{Avg. Actions} \\
        \midrule

        \textsc{Kinship} & Single
        & 112 & 248 & 420 & -- & 780
        & \num[round-mode=places, round-precision=2, scientific-notation=true]{9.1447e-02}
        & \numstats[sample][round-mode=places, round-precision=2, scientific-notation=true]{0.089208,0.093762,0.089010}
        & \num[round-mode=places, round-precision=2]{6.4189e+00} \\

        \textsc{MetaQA} & Multi
        & 116{,}045 & 148{,}724 & 142{,}744 & -- & 407{,}513
        & \num[round-mode=places, round-precision=2, scientific-notation=true]{6.8742e-02}
        & \numstats[sample][round-mode=places, round-precision=2, scientific-notation=true]{0.06867623359680763,0.06883355076356383,0.06876806589414985}
        & * \\

        \textsc{MQuAKE-ST} & Single
        & 13{,}733 & 9{,}432 & 4{,}215 & 1{,}701 & 29{,}081
        & \num[round-mode=places, round-precision=2, scientific-notation=true]{5.6292e-03}
        & \numstats[sample][round-mode=places, round-precision=2, scientific-notation=true]{0.005685,0.005545,0.005652}
        & \num[round-mode=places, round-precision=2]{4.5048e+01} \\

        \textsc{MQuAKE-ST} & Multi
        & 12{,}959 & 4{,}723 & 2{,}418 & 1{,}879 & 21{,}979
        & \num[round-mode=places, round-precision=2, scientific-notation=true]{6.0792e-03}
        & \numstats[sample][round-mode=places, round-precision=2, scientific-notation=true]{0.005874,0.006046,0.006218}
        & \num[round-mode=places, round-precision=2]{8.3360e+01} \\

        \bottomrule
    \end{tabular}
    }
    \footnotesize
    $*$ indicates that the metric is not available for this setting.
\end{table}

%
To support \algo, we build upon two existing resources, \textsc{Kinship}~\cite{kinship_55} and \textsc{MQuAKE-CF}~\cite{zhong2023mquake}, and augment them into \emph{navigation-ready} and \emph{interpretable} KGQA datasets.
%
%
Summary statistics for \textsc{Kinship} and \textsc{MQuAKE-ST} appear in Tabs.~\ref{tab:kg_stats} and~\ref{tab:qa_stats}.
For reference, we also include analogous statistics for \textsc{MetaQA}, a widely used multi-hop KGQA benchmark, to contextualize the scale and hop structure of our navigation-ready resources.
Beyond entity, relation, triple, and question counts, we report structural graph measures and random-walk baselines that characterize the navigation space, including graph density, clustering coefficient, weakly connected components (WCCs), exact random-walk answer reachability (RW-Ans$_p$), its Monte Carlo estimate (RW-Ans$_{\mathrm{MC}}$), and the average number of available actions along annotated paths.
Definitions are provided in Appendix~\ref{app:structural_metrics}.

\noindent
\textbf{\textsc{Kinship}}\footnote{\url{https://storage.googleapis.com/halcyon_data/multihop_ds/datasets/Kinship/index.html}}:
%
A toy KGQA resource derived from the UCI Kinship dataset~\cite{kinship_55}\footnote{Distinct from the Australian Kinship dataset~\cite{denham1979aranda}, which uses anonymized entities and relations.}.
The original resource provides a KG composed of two family subgraphs, with fully labeled entities and kinship relations.
We leverage this hierarchical structure to generate 1--3-hop questions by enumerating simple acyclic paths $A\!\to\!\cdots\!\to\!B$, filtering duplicates, and discarding paths that reverse direction across generational levels.
Each path is verbalized into a canonical question using a fixed compositional template and augmented with three randomly sampled paraphrase variants.
For QA splits, we perform a random 70/15/15 train/dev/test split independently for each hop size, except for 1-hop questions, which are reserved entirely for training to avoid leakage with KGC-style embedding training.
The dataset serves as a lightweight sandbox for evaluating \emph{model capacity}: effective navigation must be learned under a constrained representational budget with compact KG embeddings (i.e., $d_{KG}{=}12$).
Additional details on relation types, question templates, paraphrase variants, and sample questions are provided in Appendix~\ref{app:kinship_extra}.

\noindent
\textbf{\textsc{MQuAKE-ST}}\footnote{\url{https://storage.googleapis.com/halcyon_data/multihop_ds/datasets/MQuAKE_ST/index.html}}:
%
We build \textsc{MQuAKE-ST} from \textsc{MQuAKE-CF}~\cite{zhong2023mquake}, which already provides multi-hop questions, question variants, annotated reasoning paths, and mappings from Wikidata~\cite{vrandevcic2014wikidata} identifiers to human-readable labels for entities and relations, i.e., QIDs and PIDs, respectively.
However, \textsc{MQuAKE-CF} was designed as a general-knowledge benchmark for LLM-based knowledge editing rather than graph navigation, and is therefore not directly navigation-ready.
In particular, constructing a KG only from triples appearing in the annotated paths would yield many isolated subgraphs, restricting navigation to small isolated components.
At the QA layer, prior analysis by \cite{zhong2025mquake} report that some question texts omit information needed to specify the underlying triple chain.
We further observe duplicate annotated paths paired with different question texts, as well as snapshot-dependent factual inconsistencies when adapting the resource to a static KG.

To obtain a fixed and challenging navigation space, we reconstruct a static KG by querying Wikidata for triples in which entities from the annotated paths appear as either head or tail entities, then prune infrequent entities and relations while exempting the original seed entities and relations.
This expands the initial seed set from roughly $9$K triples, $9$K entities, and $37$ relation types to a final KG with roughly $700$K triples, $39$K entities, and $665$ relation types, as reported in Tab.~\ref{tab:kg_stats}.
We then reconstruct the QA layer using verified relation-chain templates to replace the original question texts.
For the original unusable paths, we sample replacement paths from the reconstructed KG that are compatible with the corresponding relation-chain template, retaining only those that are contiguous, present in the KG, and unique at the Wikidata-identifier level.
The resulting resource, \textsc{MQuAKE-ST}, where \emph{ST} denotes ``Static'', provides two evaluation settings: \emph{Single-Answer} and \emph{Multi-Answer}.
In the Single-Answer setting, each topic entity and relation chain is constrained to yield a single semantically valid answer, whereas in the Multi-Answer setting, all semantically valid answers are retained.
For QA splits, we perform a random 80/10/10 train/dev/test split, except for 1-hop questions, all of which we keep in the training split.
Additional construction details, released metadata, and sample questions are provided in Appendix~\ref{app:mquake_st_extra}.

\section{Experiments}
\label{sec:exp}

\begin{table*}
    \centering
    \caption{
    Main performance results across \textsc{Kinship} and \textsc{MQuAKE-ST} ($n$-hop).
    Mean $\pm$ std over three trials. 
    The answer-oracle shortest-path reference is deterministic and is reported as a single value.
    Higher is better for MRR, Hits@1, $\mathrm{F1_{Rel}}$, and $\mathrm{F1_{SG}}$; 
    lower is better for RED and PED.
    Best valid results are bolded within each model group and dataset setting.
    }
    \label{tab:combined_metric}

\resizebox{\textwidth}{!}{
    \begin{tabular}{l ccc cccc}
        

        \toprule
        \multirow{2}{*}{\textbf{Model / Metric}}
         & \multicolumn{2}{c}{\textbf{Answer Metrics}}
         & \multicolumn{4}{c}{\textbf{Path-Fidelity Metrics}} \\
         \cmidrule(lr){2-3}
         \cmidrule(lr){4-7}
        
        & MRR~$\uparrow$ & Hits@1~$\uparrow$ & $\mathrm{RED}\downarrow$  & $\mathrm{PED}\downarrow$  & $\mathrm{F1_{Rel}}\uparrow$ & $\mathrm{F1_{SG}}\uparrow$ \\
        \midrule

        
        \multicolumn{7}{l}{(a) \textsc{Kinship}, $N=3$} \\
        \toprule

        \footnotesize\textbf{Structural Calibration References} & & & & & & \\
        
        $\textsc{RW-Ans}_{\mathrm{MC}}$
        & $\dagger$ & $\dagger$ & \numstats{2.397822,2.398020,2.408218} & \numstats{2.540594,2.547426,2.550891} & \numstats{0.259713,0.259914,0.256205} & \numstats{0.128561,0.127360,0.127106} \\
        
        \textsc{Shortest Path Oracle}
        & $\dagger$ & $\dagger$ & 1.871 & 2.099 & 0.350 & 0.220 \\

        \midrule

        \footnotesize\textbf{Direct Answer-Prediction Models} & & & & & & \\

        \textsc{EmbedKGQA} 
        & \best{\numstats{0.995050,0.948114,0.962046}} & \best{\numstats{0.990099,0.920792,0.930693}} & * & * & * & * \\

        \textsc{TransferNet} 
        & \numstats{0.9033,0.9241,0.9662} & \numstats{0.8416,0.9010,0.9406} & * & * & * & * \\

        \textsc{ReaRev} 
        & \numstats{0.9389,0.8243,0.9802} & \numstats{0.9001,0.7129,0.9604} & * & * & * & * \\
        
        \midrule

        \footnotesize\textbf{Adapted Path-based Models} & & & & & & \\

        \textsc{MINERVA}
        & \best{\numstats{0.9632,0.9777,0.9571}} & \best{\numstats{0.9406,0.9604,0.9208}} & \numstats{1.8317,1.7129,2.1386} & \numstats{2.0990,1.9901,2.4257} & \numstats{0.4587,0.4818,0.3802} & \numstats{0.3129,0.3330,0.1967} \\

        \textsc{MultiHopKG} 
        & \numstats{0.9273,0.9686,0.9653} & \numstats{0.8812,0.9406,0.9307} & \numstats{2.3366,2.1089,2.3267} & \numstats{2.4752,2.2673,2.5050} & \numstats{0.2422,0.3281,0.2488} & \numstats{0.1416,0.2300,0.1191} \\

        \textsc{SQUIRE} 
        & \numstats{0.584512,0.654258,0.555333} & \numstats{0.386139,0.490000,0.350000} & \best{\numstats{1.188119,1.150000,1.470000}} & \best{\numstats{1.643564,1.610000,1.820000}} & \best{\numstats{0.578307,0.644667,0.592333}} & \best{\numstats{0.370370,0.429000,0.366667}} \\

        
        \midrule
        \multicolumn{7}{l}{(b) \textsc{MQuAKE-ST Single Answer}, $N=4$} \\

        \toprule

        \footnotesize\textbf{Structural Calibration References} & & & & & & \\
        
        $\textsc{RW-Ans}_{\mathrm{MC}}$
        & $\dagger$ & $\dagger$ & \numstats{3.343597,3.340944,3.347267} & \numstats{3.379761,3.378072,3.383730} & \numstats{0.056511,0.055794,0.055651} & \numstats{0.034051,0.033324,0.033369} \\
        
        \textsc{Shortest Path Oracle}
        & $\dagger$ & $\dagger$ & 1.418 & 1.670 & 0.503 & 0.380 \\

        \midrule

        \footnotesize\textbf{Direct Answer-Prediction Models} & & & & & & \\

        \textsc{EmbedKGQA} 
        & \numstats{0.702788,0.683850,0.680427} & \numstats{0.567819,0.563830,0.557846} & * & * & * & * \\

        \textsc{TransferNet} 
        & \numstats{0.6600,0.5282,0.5285} & \numstats{0.6117,0.4847,0.4847} & * & * & * & * \\

        \textsc{ReaRev}
        & \best{\numstats{0.9863,0.9874,0.9848}} & \best{\numstats{0.9840,0.9847,0.9827}} & * & * & * & * \\
        \midrule

        \footnotesize\textbf{Adapted Path-based Models} & & & & & & \\
        
        \textsc{MINERVA}
        & \best{\numstats{0.9263,0.9221,0.9019}} & \best{\numstats{0.8916,0.8863,0.8610}} & \numstats{0.5372,0.6602,0.5652} &
        \best{\numstats{0.7786,0.9215,0.7932}} & \numstats{0.8393,0.8091,0.8264} & \best{\numstats{0.7332,0.6960,0.7260}} \\ 

        \textsc{MultiHopKG} 
        & \numstats{0.7740,0.7675,0.7928} & \numstats{0.7141,0.7134,0.7387} & \numstats{2.6449,2.6882,2.5439} & \numstats{2.9428,2.9953,2.9049} & \numstats{0.3359,0.3286,0.3638} & \numstats{0.1930,0.1920,0.2092} \\

        \textsc{SQUIRE} 
        & \numstats{0.692027,0.702873,0.712830} & \numstats{0.592420,0.603940,0.614810} & \best{\numstats{0.494681,0.429348,0.437500}} & \numstats{1.337766,1.315897,1.311141} & \best{\numstats{0.826732,0.850684,0.845846}} & \numstats{0.455107,0.464742,0.466073} \\

        
        \midrule
        \multicolumn{7}{l}{(c) \textsc{MQuAKE-ST Multi Answer}, $N=4$} \\
       
        \toprule

        \footnotesize\textbf{Structural Calibration References} & & & & & & \\

        $\textsc{RW-Ans}_{\mathrm{MC}}$
        & $\dagger$ & $\dagger$ & \numstats{3.368805,3.370839,3.369713} & \numstats{3.404241,3.406517,3.405977} & \numstats{0.066682,0.066032,0.066520} & \numstats{0.044360,0.043692,0.044209} \\
        
        \textsc{Shortest Path Oracle}
        & $\dagger$ & $\dagger$ & 1.702 & 1.843 & 0.437 & 0.362 \\

        \midrule

        \footnotesize\textbf{Direct Answer-Prediction Models} & & & & & & \\

        \textsc{EmbedKGQA} 
        & \numstats{0.903772,0.912541,0.909349} & \numstats{0.868966,0.878161,0.872414} & * & * & * & * \\

        \textsc{TransferNet} 
        & \numstats{0.6461,0.4650,0.8324} & \numstats{0.5954,0.4299,0.8092} & * & * & * & * \\

        \textsc{ReaRev}
        & \best{\numstats{0.9763,0.9788,0.9691}} & \best{\numstats{0.9643,0.9678,0.9552}} & * & * & * & * \\

        \midrule

        \footnotesize\textbf{Adapted Path-based Models} & & & & & & \\
        
        \textsc{MINERVA}
        & \best{\numstats{0.9283,0.8891,0.9093}} & \best{\numstats{0.8966,0.8563,0.8701}} & \numstats{1.0690,1.2023,1.3644} & \numstats{1.3057, 1.4264,1.5828} & \numstats{0.7181,0.6449,0.6106} & \best{\numstats{0.6174, 0.5451,0.5047}} \\

        \textsc{MultiHopKG} 
        & \numstats{0.8930,0.9018,0.8876} & \numstats{0.8506,0.8598,0.8425} & \numstats{2.1828,2.2080,2.1759} & \numstats{2.4437,2.4368,2.4046} &  \numstats{0.3685,0.3630,0.3690} & \numstats{0.2661,0.2606,0.2707} \\

        \textsc{SQUIRE} 
        & \numstats{0.738829,0.761474,0.761474} & \numstats{0.636782,0.621839,0.667816} & \best{\numstats{0.449425,0.531034,0.495402}} & \best{\numstats{1.337766,1.317819,1.312500}} & \best{\numstats{0.858131,0.832915,0.840099}} & \numstats{0.455107,0.463442,0.465077} \\
        \bottomrule

    \end{tabular}
    }

    \footnotesize\emph{
    $*$: Not applicable; method does not expose path-level predictions.
    $\dagger$: Not applicable; structural calibration reference does not produce a ranked answer prediction, so MRR and Hits@1 are undefined.
    %
    }
\end{table*} 

\noindent
\textbf{Protocol:}
We evaluate using the candidate-based answer and path-ranking procedure defined in Sec.~\ref{sec:problem_formulation} and Appendix~\ref{app:rollout_ranking}.
For the reported experiments, all adapted navigation methods use beam search with a common candidate budget of $B_{\mathrm{roll}}=100$ per question; beam search is used only as an inference procedure and does not alter their respective training objectives.
This fixes a common candidate budget across methods; for sampling-based methods, the larger candidate set also reduces variability due to stochastic candidate generation.
\textsc{MINERVA} and \textsc{MultiHopKG} score candidate trajectories by cumulative policy log-probability, whereas \textsc{SQUIRE} uses cumulative sequence log-probability.
Each episode runs for exactly $N$ decision steps, where $N$ is set to the maximum annotated hop length in the corresponding dataset.
We treat the KG as directed for both \textsc{Kinship} and \textsc{MQuAKE-ST}.
In the Multi-Answer setting, any terminal entity in $\mathcal{A}(q)$ is treated as a valid answer for answer ranking.
For the adapted navigation baselines, we use a frozen BERT-base-uncased encoder~\cite{devlin2019bert} and instantiate $\phi(\cdot)$ as a linear projection, yielding a simple yet effective mechanism for conditioning navigation on the question.
Unless otherwise stated, KG embedding dimensions are set to $d_{\mathrm{KG}}=12$ for \textsc{Kinship}, reflecting its toy scale, and $d_{\mathrm{KG}}=100$ for \textsc{MQuAKE-ST}.
We run three trials with different random seeds and report the mean and standard deviation across trials.

\noindent
\textbf{Baselines.}
To calibrate the scale of the path-fidelity metrics, we include two non-learned structural references that are not conditioned on the question: the same unbiased random-walk baseline used for RW-Ans$_{\mathrm{MC}}$ and the \textsc{Shortest Path Oracle}, the latter of which is given the valid answer set.
These references serve as structural calibration points rather than competing KGQA methods; the full definitions are provided in Appendix~\ref{app:structural_metrics}.
In addition to the adapted path-based navigation models, we include direct answer-prediction baselines from the KGQA literature: \textsc{EmbedKGQA}, \textsc{TransferNet}, and \textsc{ReaRev}.
These models are trained from scratch using the same dataset splits and KG preprocessing whenever applicable.
They are included to contextualize answer-prediction performance rather than as direct path-level comparisons to navigation agents, since they do not explicitly navigate the KG or produce complete trajectories.
For answer evaluation, the direct models score candidate entities directly, whereas the navigation models derive an entity ranking from the terminal entities of their scored candidate trajectories.
For the direct models, the candidate space is the full KG for \textsc{EmbedKGQA} and \textsc{TransferNet}, and the question-specific subgraph for \textsc{ReaRev}.
%
%
Additional implementation and reproducibility details are provided in Appendix~\ref{app:reproducibility}.

We analyze PQ, PQL, and WC2014 separately in Appendix~\ref{app:prior_dataset} as structural reference benchmarks rather than additional model-evaluation datasets, due to differences in their graph construction and evaluation protocols.
Although IRN and SRN are closely related path-explicit KGQA models, we do not include them in the main empirical comparison due to reproducibility barriers in the available implementations and release artifacts.
We provide details in Appendix~\ref{app:srn_irn_reproducibility}.

\begin{table*}
    \caption{
    Hits@1 and path edit distance (PED) performance under question rephrasing.
    Models are trained on the canonical questions and evaluated on their rephrased variants, using the settings reported in Tab.~\ref{tab:combined_metric}.
    Higher is better for Hits@1; lower is better for PED.
    %
    Best valid results are bolded within each model group.
    }
    \label{tab:paraphrased_metrics}
    \resizebox{\textwidth}{!}{
    \begin{tabular}{l cc cc cc}

        \toprule
        \multirow{2}{*}{\textbf{Model / Metric}}
        & \multicolumn{2}{c}{\textbf{\textsc{Kinship}}}
        & \multicolumn{2}{c}{\textbf{\textsc{MQuAKE-ST} Single}}
        & \multicolumn{2}{c}{\textbf{\textsc{MQuAKE-ST} Multi}} \\
        \cmidrule(lr){2-3}
        \cmidrule(lr){4-5}
        \cmidrule(lr){6-7}

        & Hits@1~$\uparrow$ & PED~$\downarrow$
        & Hits@1~$\uparrow$ & PED~$\downarrow$
        & Hits@1~$\uparrow$ & PED~$\downarrow$ \\
        \midrule


        \footnotesize\textbf{Direct Answer-Prediction Models}
        & & & & & & \\

        \textsc{EmbedKGQA} 
        & \numstats{0.132013,0.194719,0.194719} & *
        & \numstats{0.564495,0.562722,0.554743} & *
        & \numstats{0.847893,0.849425,0.842529} & * \\
        
        \textsc{TransferNet} 
        & \best{\numstats{0.3135,0.4950,0.4752}} & * 
        & \numstats{0.3260,0.5536,0.4488} & * 
        & \numstats{0.5716,0.4395,0.7992} & * \\

        \textsc{ReaRev} 
        & \numstats{0.2145,0.2937,0.2871} & *
        & \best{\numstats{0.9515,0.9579,0.9508}} & *
        & \best{\numstats{0.9349,0.9471,0.9303}} & * \\

        \midrule


        \footnotesize\textbf{Adapted Path-based Models}
        & & & & & & \\
        
        \textsc{MINERVA} 
        & \numstats{0.2277,0.2112,0.1716} & \numstats{2.5215, 2.3762, 2.5908}
        & \best{\numstats{0.8719,0.8630,0.8469}} & \numstats{0.7934, 0.9424, 0.8025}
        & \best{\numstats{0.8567,0.8238,0.8433}} & \numstats{1.3625, 1.4395, 1.5931} \\
        
        \textsc{MultiHopKG} 
        & \numstats{0.2409,0.2079,0.2475} & \numstats{2.5842,2.5512,2.6304}
        & \numstats{0.7077,0.6990,0.7323} & \numstats{2.9399,2.9958,2.8985}
        & \numstats{0.8464,0.8437,0.8272} & \numstats{2.4356,2.4157,2.4000} \\
        
        \textsc{SQUIRE} 
        & \best{\numstats{0.280000,0.203333,0.280000}} & \numstats{2.247525,2.198020,2.250825} 
        & \numstats{0.586879,0.601950,0.607491} & \numstats{1.337766,1.318706,1.310284} 
        & \numstats{0.640230,0.617625,0.670115} & \numstats{1.418391,1.458238,1.363985} \\
        
        \bottomrule
        
      \end{tabular}
      }
\end{table*}

\noindent
\textbf{Main Results.}
Table~\ref{tab:combined_metric} reports the main $n$-hop results on \textsc{Kinship} and \textsc{MQuAKE-ST}.
The structural calibration references provide complementary context for interpreting the path-fidelity metrics: the unbiased random-walk baseline underlying RW-Ans$_{\mathrm{MC}}$ characterizes unguided traversal, whereas the \textsc{Shortest Path Oracle} characterizes an answer-informed but question-agnostic route that prioritizes graph-theoretic efficiency.
The latter is not a strict upper or lower bound on path fidelity, since the shortest route to a valid answer need not coincide with the reasoning path implied by the question.
Instead, it provides a structural reference for how the path-fidelity metrics behave when answer reachability is guaranteed but the question is not used to select the route.
Accordingly, outperforming this reference on a path-fidelity metric indicates closer agreement with the annotated reference reasoning structure under that metric than is obtained from answer knowledge and shortest-path efficiency alone, whereas performance approaching the random-walk reference indicates increasingly weak trajectory alignment.
The direct answer-prediction models provide complementary answer-ranking context, with \textsc{ReaRev} performing particularly strongly on \textsc{MQuAKE-ST}.

On \textsc{Kinship}, \textsc{MINERVA} achieves the strongest answer-ranking performance among the path-based agents.
Its path fidelity is also substantially stronger than that of \textsc{MultiHopKG}, whose path metrics lie much closer to the random-walk calibration and remain weaker than the shortest-path oracle.
This contrast shows that strong answer ranking does not by itself imply close agreement between the executed trajectory and the reference evidence path.
\textsc{SQUIRE} exhibits the opposite profile: despite substantially weaker answer ranking, it achieves the strongest path fidelity across all four path metrics and outperforms the shortest-path oracle on each of them.
The \textsc{Kinship} results therefore expose a clear separation between answer-reaching performance and trajectory fidelity.

On \textsc{MQuAKE-ST}, the path-based models again exhibit complementary strengths.
\textsc{MINERVA} provides the strongest answer-ranking performance among the path-based agents and, in the single-answer setting, the strongest entity-level path fidelity.
\textsc{SQUIRE}, in contrast, provides the strongest relation-level fidelity in both answer settings.
In the multi-answer setting, the complete entity-level evaluations of \textsc{MINERVA} and \textsc{SQUIRE} further distinguish the two notions of path agreement: \textsc{SQUIRE} achieves the strongest PED, whereas \textsc{MINERVA} achieves the strongest $\mathrm{F1_{SG}}$.
This difference illustrates that ordered agreement with a reference path and order-invariant edge overlap capture distinct aspects of trajectory fidelity.
Both \textsc{MINERVA} and \textsc{SQUIRE} outperform the shortest-path oracle across the reported path-fidelity metrics in both \textsc{MQuAKE-ST} settings, indicating closer agreement with the annotated reference reasoning structure than is provided by the answer-informed structural reference.

\textsc{MultiHopKG} shows a different trade-off.
In the single-answer setting, its answer-ranking and path-fidelity performance are both weaker than those of \textsc{MINERVA}, with its path metrics generally falling between the random-walk and shortest-path calibration references.
In the multi-answer setting, its answer ranking improves substantially and approaches that of \textsc{MINERVA}, while its relation-level fidelity remains weaker than that of \textsc{MINERVA} and \textsc{SQUIRE}.
Its entity-level fidelity remains comparatively weak: PED remains well above the shortest-path oracle and the results of \textsc{MINERVA} and \textsc{SQUIRE}, while $\mathrm{F1_{SG}}$, although substantially above the random-walk reference, remains below the shortest-path oracle.
Taken together, the results show that answer correctness, relation-level fidelity, and entity-level path fidelity capture complementary properties of graph reasoning: an agent can rank a valid answer highly while differing substantially in how closely its executed trajectory matches the annotated reference reasoning structure.

Table~\ref{tab:paraphrased_metrics} reports robustness under question rephrasing.
For \textsc{Kinship}, the canonical questions are generated from a single template family, so the rephrased variants introduce a substantial linguistic distribution shift.
Answer-ranking performance consequently degrades markedly across model families, although \textsc{TransferNet} degrades less severely than the other direct answer-prediction models.
The path-based agents exhibit a corresponding deterioration in trajectory fidelity: their PED values shift toward the random-walk calibration, with \textsc{MultiHopKG} reaching approximately unguided-traversal levels, while \textsc{SQUIRE} remains between the shortest-path and random-walk references.
Together, these changes suggest that the linguistic shift substantially weakens effective question-conditioned guidance on \textsc{Kinship}.

For \textsc{MQuAKE-ST}, the models are substantially more robust to rephrasing.
Answer-ranking performance changes only modestly for the strongest models, and the single-answer PED results remain similarly stable across the path-based agents.
In the multi-answer setting, the adapted path-based agents likewise show only modest changes in PED, indicating that their entity-level trajectory fidelity is largely preserved under paraphrasing.
Overall, the stability of both answer ranking and the observed path-fidelity results indicates that the navigation policies retain substantially more of their navigation performance under rephrasing on \textsc{MQuAKE-ST} than on \textsc{Kinship}.

Additional per-hop and \textsc{MetaQA} simulations are reported in Appendix~\ref{app:additional_simulations}.

\paragraph{Limitations.}
Our setting assumes that the topic entity is already identified and that a suitable KG is available for navigation.
Although we construct curated KGs for our datasets, we do not address automatic KG construction or entity linking.
In addition, our datasets focus on questions that can be answered by sequential navigation from a single topic entity along one or more relational paths.
This excludes more complex structures that arise in natural-language questions, such as multiple topic entities, conjunctions, negation, comparison, or other logical operations.
The question templates also require manual verification to ensure that the generated questions preserve the intended relation semantics.
As a result, the number of linguistic variations and relation patterns covered by the datasets is necessarily limited.
Extending question-conditioned graph navigation to more diverse question forms, broader relation semantics, and more compositional query types is an important direction for future work.
Finally, the annotated paths should be interpreted as reference evidence paths rather than necessarily unique explanations.
Although our multi-answer evaluation accounts for alternative entity-level paths consistent with the annotated relation-chain semantics, other semantically valid reasoning routes may exist.
Accordingly, PED, RED, F1$_{\mathrm{SG}}$, and F1$_{\mathrm{Rel}}$ measure agreement with the available valid reference paths, rather than establishing that these paths are the unique correct explanations.

\section{Conclusion}
\label{sec:conclusion}

We introduced \textsc{Theseus}, a formulation of multi-hop KGQA as question-conditioned graph navigation in which both the reached answer and the executed reasoning trajectory are explicit evaluation targets.
To support this setting, we constructed navigation-ready versions of \textsc{Kinship} and \textsc{MQuAKE-ST}, developed complementary answer- and path-level evaluation protocols with structural calibration references, and adapted representative path-based reasoning models to condition their traversal on natural-language questions.

Our experiments show that answer correctness alone provides an incomplete view of graph reasoning quality.
Across models and datasets, strong answer ranking can coexist with substantially different levels of relation- and entity-level trajectory fidelity, while models that more closely reproduce the intended reasoning structure need not provide the strongest answer ranking.
The rephrasing experiments further show that question-conditioned navigation can vary considerably in its robustness to linguistic variation, highlighting the importance of evaluating not only whether an agent reaches a valid answer, but also how it reaches that answer.
Taken together, these findings motivate treating the reasoning trajectory as a first-class output of multi-hop KGQA rather than as an incidental by-product of answer prediction.
We hope that the \textsc{Theseus} setting, its datasets, and its evaluation protocols provide a foundation for developing KGQA systems whose answers are accompanied by reasoning paths that can be inspected, compared, and systematically evaluated.

\section*{Acknowledgment}
This work is partially funded by the NSTC grant number 113-2923-E-A49-001 and by MARC, the MediaTek Advanced Research Center with grant number 114A540531.

{
\small
\bibliographystyle{abbrvnat} 
\bibliography{reference}
}
\newpage
\appendix

\section{Additional Details for Problem Formulation}
\label{app:problem_formulation_details}

\subsection{KG Embedding Scoring Functions}
\label{app:kg_embedding_scoring}

Let $d_{\mathrm{KG}}$ be the embedding dimension, and let the entity and relation
embedding matrices be
$\mathbf{E}\in\mathbb{R}^{|\mathcal{E}|\times d_{\mathrm{KG}}}$ and
$\mathbf{R}\in\mathbb{R}^{|\mathcal{R}|\times d_{\mathrm{KG}}}$.
For arbitrary entities $u,v\in\mathcal{E}$ and relation $r\in\mathcal{R}$, denote
their embeddings by
$\mathbf{e}_u:=\mathbf{E}_{[u,:]}$,
$\mathbf{e}_v:=\mathbf{E}_{[v,:]}$, and
$\mathbf{r}:=\mathbf{R}_{[r,:]}$.
A triple $(u,r,v)$ is scored by
\begin{equation}
    s(u,r,v)=g \big(\mathbf{e}_u,\mathbf{r},\mathbf{e}_v\big),
\end{equation}
where $g$ is a KG scoring function.
Unless otherwise specified, embeddings are real-valued.
Common choices include the following:

\noindent $\bullet$ \emph{TransE} -- \cite{bordes2013translating}:
\begin{equation}
    s_{\text{TransE}}(u,r,v)
    =
    -\|\mathbf{e}_u+\mathbf{r}-\mathbf{e}_v\|_{p},
    \qquad p\in\{1,2\}.
\end{equation}

\noindent $\bullet$ \emph{DistMult} -- \cite{yang2015embedding}:
\begin{equation}
    s_{\text{DistMult}}(u,r,v)
    =
    \langle \mathbf{e}_u,\mathbf{r},\mathbf{e}_v\rangle,
\end{equation}
where
$\langle \mathbf{a},\mathbf{b},\mathbf{c}\rangle
:=\sum_k a_k b_k c_k$,
which is symmetric in the entity arguments $u$ and $v$.

\noindent $\bullet$ \emph{ComplEx} -- \cite{trouillon2016complex}:
\begin{equation}
    s_{\text{ComplEx}}(u,r,v)
    =
    \Re\!\big(
    \langle \mathbf{e}_u,\mathbf{r},\overline{\mathbf{e}_v}\rangle
    \big),
    \qquad
    \mathbf{e}_u,\mathbf{r},\mathbf{e}_v\in\mathbb{C}^{d_{\mathrm{KG}}},
\end{equation}
where $\Re(\cdot)$ denotes the real part and $\overline{\mathbf{e}_v}$ denotes complex conjugation.

\noindent $\bullet$ \emph{RotatE} -- \cite{sun2019rotate}:
\begin{equation}
    s_{\text{RotatE}}(u,r,v)
    =
    -\|\mathbf{e}_u\circ\mathbf{r}-\mathbf{e}_v\|_{1},
    \qquad
    \mathbf{e}_u,\mathbf{r},\mathbf{e}_v\in\mathbb{C}^{d_{\mathrm{KG}}},
    \quad |[\mathbf{r}]_k|=1,
\end{equation}
where $\circ$ denotes element-wise multiplication.

\noindent $\bullet$ \emph{ConvE} -- \cite{dettmers2018convolutional}:
\begin{equation}
    s_{\text{ConvE}}(u,r,v)
    =
    \eta\!\left(
    \operatorname{vec}\!\left(
    \eta\!\left(
    [\widetilde{\mathbf{e}}_u;\widetilde{\mathbf{r}}]
    \ast\boldsymbol{\Omega}
    \right)\right)\mathbf{W}
    \right)^{\top}\mathbf{e}_v,
\end{equation}
where $\widetilde{\mathbf{e}}_u$ and $\widetilde{\mathbf{r}}$ are two-dimensional reshapes of the embeddings, $\boldsymbol{\Omega}$ and $\mathbf{W}$ are learned convolutional and projection parameters, $\ast$ denotes convolution, and $\eta$ is a nonlinear activation.

These scores provide structural priors for ranking candidate outgoing actions $(r,e')$ from a current entity $u$.
They are typically learned via link prediction, i.e., by training $g$ to assign higher scores to observed triples than to corrupted triples.

\subsection{Candidate-Based Answer and Path Ranking}
\label{app:rollout_ranking}

For each question $(q,e_s)$, we generate $B_{\mathrm{roll}}$ candidate trajectories using the candidate-generation procedure of the evaluated method, such as beam search or independent policy sampling.
No particular search procedure is required; the evaluation only assumes that the method produces a scored set of candidate trajectories.
Each candidate trajectory $\tau^{(j)}$ is assigned a model score $s^{(j)}$, such as its sequence or policy log-probability, and the candidates are ordered by decreasing score.
Let $\rho$ denote the resulting ordering:
\[
    s^{(\rho_1)}
    \ge
    s^{(\rho_2)}
    \ge
    \cdots
    \ge
    s^{(\rho_{B_{\mathrm{roll}}})}.
\]

\paragraph{Answer ranking.}
Multiple trajectories may terminate at the same entity.
To obtain an answer ranking comparable to entity-ranking KGQA methods, we collapse such duplicates after trajectory scoring.
For each terminal entity $e$, we retain the score of its highest-ranked trajectory,
\begin{equation}
    S(e)
    =
    \max_{j:\,\term(\tau^{(j)})=e}
    s^{(j)}.
\end{equation}
The unique terminal entities are then ranked in decreasing order of $S(e)$.
Let
\[
    (e_{(1)},e_{(2)},\ldots,e_{(M)}),
    \qquad M\le B_{\mathrm{roll}},
\]
denote this ranked list.
The answer rank is
\begin{equation}
    \rankf(q)=
    \begin{cases}
        \displaystyle
        \min\left\{
            k\in\{1,\ldots,M\}:
            e_{(k)}\in\mathcal{A}(q)
        \right\},
        & \text{if a valid answer is retrieved},\\[1ex]
        B_{\mathrm{roll}}+1,
        & \text{otherwise}.
    \end{cases}
\end{equation}
Thus, questions for which no candidate reaches a valid answer receive the fixed worst rank $B_{\mathrm{roll}}+1$.
Hits@$K$ indicates whether any valid answer occurs among the top-$K$ unique terminal entities, while MRR is the reciprocal rank of the highest-ranked valid answer entity, with zero contribution when no valid answer is retrieved.

\paragraph{Path-fidelity evaluation.}
Path-fidelity metrics are evaluated separately from the deduplicated answer ranking.
We use the single highest-scoring trajectory
\begin{equation}
    \widehat{\tau}(q)
    =
    \tau^{(\rho_1)}
\end{equation}
and compare its induced path against the valid reference evidence path(s) using the path-fidelity metrics defined in Appendix~\ref{app:path_fidelity_metrics}.
Thus, answer metrics evaluate the ranking of unique candidate answer entities, whereas path-fidelity metrics evaluate the reasoning trajectory preferred most strongly by the model.

\subsection{Path-Fidelity Metrics}
\label{app:path_fidelity_metrics}

When annotated reasoning paths are available, we evaluate the structural agreement between the model's top-ranked executed trajectory and the reference reasoning structure, independently of whether its terminal entity is a valid answer.
These metrics are diagnostic and are computed only on the top-ranked trajectory $\widehat{\tau}(q)$, i.e., the candidate trajectory assigned the highest model score.

Before computing path-fidelity metrics, implementation-specific control actions such as \textsc{No Operation} are removed from the predicted trajectory.
Let $\Psf^\star(q)$ denote the annotated reference path from Eq.~\eqref{eq:path}, and let $\widehat{\Psf}(q)$ denote the path induced by the top-ranked trajectory $\widehat{\tau}(q)$.

\paragraph{Path edit distance.}
The path edit distance compares the ordered edge sequence of the predicted path against the ordered edge sequence of the annotated reference path:
\begin{equation}
    \mathrm{PED}(q)
    =
    \Lev\!\left(
    \widehat{\Psf}(q),
    \Psf^\star(q)
    \right),
\end{equation}
where $\Lev(\cdot,\cdot)$ denotes the Levenshtein edit distance between ordered edge sequences.
Lower values indicate closer agreement with the annotated reasoning path, with zero corresponding to an exact match.

\paragraph{Relation edit distance.}
We also compute relation edit distance by comparing only the ordered sequence of relations in each path.
Using the relation-sequence operator $\Rsf(\cdot)$ from Eq.~\eqref{eq:relation_sequence}, the relation edit distance is
\begin{equation}
    \mathrm{RED}(q)
    =
    \Lev\!\left(
    \Rsf(\widehat{\Psf}(q)),
    \Rsf(\Psf^\star(q))
    \right).
\end{equation}
Unlike $\mathrm{PED}$, which compares complete triples, $\mathrm{RED}$ ignores the intermediate entities and measures whether the predicted path follows the same relational composition.
This makes it more tolerant in multi-answer settings, where different answer entities may be reached through the same or similar relation pattern.

\paragraph{Subgraph-overlap F1.}
As a more relaxed path-fidelity metric, we measure order-invariant structural overlap.
Here, each path is treated as a small subgraph induced by the set of edges it traverses, rather than as a strictly ordered sequence of transitions.
This measures structural overlap even when the traversal order or complete sequence differs from the reference.

Let
\begin{equation}
    \Edges(\Psf)
    =
    \left\{
    (e_{i-1},r_i,e_i)
    :
    (e_{i-1},r_i,e_i)\in\Psf
    \right\}.
\end{equation}
The subgraph-overlap precision and recall are
\begin{equation}
    \mathrm{Prec}_{\mathrm{SG}}(q)
    =
    \begin{cases}
    \dfrac{
    |\Edges(\widehat{\Psf}(q))\cap \Edges(\Psf^\star(q))|
    }{
    |\Edges(\widehat{\Psf}(q))|
    },
    & |\Edges(\widehat{\Psf}(q))|>0,\\[1.0em]
    0, & \text{otherwise,}
    \end{cases}
\end{equation}
\begin{equation}
    \mathrm{Rec}_{\mathrm{SG}}(q)
    =
    \begin{cases}
    \dfrac{
    |\Edges(\widehat{\Psf}(q))\cap \Edges(\Psf^\star(q))|
    }{
    |\Edges(\Psf^\star(q))|
    },
    & |\Edges(\Psf^\star(q))|>0,\\[1.0em]
    0, & \text{otherwise.}
    \end{cases}
\end{equation}
The subgraph-overlap F1 score is
\begin{equation}
    \mathrm{F1}_{\mathrm{SG}}(q)
    =
    \begin{cases}
    \dfrac{
    2\,\mathrm{Prec}_{\mathrm{SG}}(q)\,\mathrm{Rec}_{\mathrm{SG}}(q)
    }{
    \mathrm{Prec}_{\mathrm{SG}}(q)+\mathrm{Rec}_{\mathrm{SG}}(q)
    },
    & \mathrm{Prec}_{\mathrm{SG}}(q)+\mathrm{Rec}_{\mathrm{SG}}(q)>0,\\[1.0em]
    0, & \text{otherwise.}
    \end{cases}
\end{equation}

\paragraph{Relation-overlap F1.}
We also compute an order-invariant relation-overlap F1 score.
While relation edit distance compares the ordered relation sequence $\Rsf(\Psf)$ from Eq.~\eqref{eq:relation_sequence}, relation-overlap F1 ignores order and compares the set of relation types used by the predicted and annotated paths.
This provides a more relaxed measure of whether the agent follows a similar relational composition, even when the exact traversal order differs.

Let
\begin{equation}
    \Relations(\Psf)
    =
    \left\{
    r_i
    :
    (e_{i-1},r_i,e_i)\in\Psf
    \right\}.
\end{equation}
The relation-overlap precision,  recall, and F1 score are defined as
\eas{
    \mathrm{Prec}_{\mathrm{Rel}}(q)
    & =
    \begin{cases}
    \dfrac{
    |\Relations(\widehat{\Psf}(q))\cap \Relations(\Psf^\star(q))|
    }{
    |\Relations(\widehat{\Psf}(q))|
    },
    & |\Relations(\widehat{\Psf}(q))|>0,\\[1.0em]
    0, & \text{otherwise,}
    \end{cases}
\\
    \mathrm{Rec}_{\mathrm{Rel}}(q)
    & =
    \begin{cases}
    \dfrac{
    |\Relations(\widehat{\Psf}(q))\cap \Relations(\Psf^\star(q))|
    }{
    |\Relations(\Psf^\star(q))|
    },
    & |\Relations(\Psf^\star(q))|>0,\\[1.0em]
    0, & \text{otherwise.}
    \end{cases}
\\
    \mathrm{F1}_{\mathrm{Rel}}(q)
   &  =
    \begin{cases}
    \dfrac{
    2\,\mathrm{Prec}_{\mathrm{Rel}}(q)\,\mathrm{Rec}_{\mathrm{Rel}}(q)
    }{
    \mathrm{Prec}_{\mathrm{Rel}}(q)+\mathrm{Rec}_{\mathrm{Rel}}(q)
    },
    & \mathrm{Prec}_{\mathrm{Rel}}(q)+\mathrm{Rec}_{\mathrm{Rel}}(q)>0,\\[1.0em]
    0, & \text{otherwise.}
    \end{cases}
}

\paragraph{Multi-answer extension.}
The definitions above consider a single annotated reference path $\Psf^\star(q)$.
In the Multi-Answer setting, however, a question may admit multiple semantically valid entity-level reference paths.
Let $\mathcal{P}^\star(q)$ denote the set of all valid reference paths for question $q$ under its annotated relation-chain semantics.
For \textsc{MQuAKE-ST}, $\mathcal{P}^\star(q)$ is reconstructed on demand during evaluation by traversing the static KG according to the annotated relation chain.
For metrics that depend on the complete entity-level path, we apply best-reference matching:
\begin{equation}
    \mathrm{PED}_{\mathrm{multi}}(q)
    =
    \min_{\Psf^\star \in \mathcal{P}^\star(q)}
    \Lev\!\left(
        \widehat{\Psf}(q),
        \Psf^\star
    \right),
\end{equation}
and
\begin{equation}
    \mathrm{F1}_{\mathrm{SG,multi}}(q)
    =
    \max_{\Psf^\star \in \mathcal{P}^\star(q)}
    \mathrm{F1}_{\mathrm{SG}}
    \!\left(
        \widehat{\Psf}(q),
        \Psf^\star
    \right),
\end{equation}
where $\mathrm{F1}_{\mathrm{SG}}(\widehat{\Psf},\Psf^\star)$ denotes the pairwise subgraph-overlap score defined above.
Thus, a prediction is not penalized for following one semantically valid entity-level realization rather than another.
In \textsc{MQuAKE-ST} Multi-Answer, all valid reference paths share the same ordered relation chain; consequently, $\mathrm{RED}$ and $\mathrm{F1}_{\mathrm{Rel}}$ are invariant to the choice of entity-level reference path and are computed against this shared relation structure.
In the Multi-Answer setting, these best-reference quantities replace $\mathrm{PED}(q)$ and $\mathrm{F1}_{\mathrm{SG}}(q)$ in all dataset-level averages and reported results.

\paragraph{Dataset-level averages.}
For the subset $\mathcal{Q}_{\mathrm{path}}\subseteq\mathcal{Q}$ with annotated reference paths, we report the average path-fidelity scores:
%
\eas{
    \mathrm{PED}
     & =
    \frac{1}{|\mathcal{Q}_{\mathrm{path}}|}
    \sum_{q\in\mathcal{Q}_{\mathrm{path}}}
    \mathrm{PED}(q) \\
    \mathrm{RED}
    & =
    \frac{1}{|\mathcal{Q}_{\mathrm{path}}|}
    \sum_{q\in\mathcal{Q}_{\mathrm{path}}}
    \mathrm{RED}(q) \\
    \mathrm{F1}_{\mathrm{SG}}
    & =
    \frac{1}{|\mathcal{Q}_{\mathrm{path}}|}
    \sum_{q\in\mathcal{Q}_{\mathrm{path}}}
    \mathrm{F1}_{\mathrm{SG}}(q) \\
    \mathrm{F1}_{\mathrm{Rel}}
    & =
    \frac{1}{|\mathcal{Q}_{\mathrm{path}}|}
    \sum_{q\in\mathcal{Q}_{\mathrm{path}}}
    \mathrm{F1}_{\mathrm{Rel}}(q).
}

\paragraph{Metric ranges.}
The edit-distance metrics are non-negative integers.
Since relation sequences are obtained by projecting each edge to its relation label, relation edit distance
cannot exceed path edit distance:
\begin{equation}
    0
    \le
    \mathrm{RED}(q)
    \le
    \mathrm{PED}(q)
    \le
    \max\left\{
    |\widehat{\Psf}(q)|,\,
    |\Psf^\star(q)|
    \right\}.
\end{equation}
Because predicted trajectories are generated with hop budget $N$ and annotated paths have length at most $N$, this further implies
\begin{equation}
    0
    \le
    \mathrm{RED}(q)
    \le
    \mathrm{PED}(q)
    \le
    N.
\end{equation}
Lower values indicate better path agreement.
In contrast, the overlap-based metrics satisfy
\begin{equation}
    0
    \le
    \mathrm{F1}_{\mathrm{Rel}}(q),
    \mathrm{F1}_{\mathrm{SG}}(q)
    \le
    1,
\end{equation}
where higher values indicate greater overlap with the annotated reasoning path.
Although $\mathrm{F1}_{\mathrm{Rel}}$ is more relaxed because it ignores intermediate entities and edge identity, it is not necessarily an upper bound on $\mathrm{F1}_{\mathrm{SG}}$ under the set-based definition, since repeated relation types are collapsed.

\paragraph{Relation to prior interpretability metrics.}
\cite{lv2021multi} evaluate multi-hop reasoning paths for KG link prediction using Path Recall (PR), Local Interpretability (LI), and Global Interpretability (GI).
PR measures whether a model recovers an answer-reaching path, while LI evaluates the reasonableness of successful paths using manually annotated rule-level interpretability scores; GI combines the two.
Thus, their framework evaluates whether answer-reaching paths in a link prediction setting are human-interpretable.
In contrast, our path-fidelity metrics do not assess human-perceived plausibility; PED, RED, F1$_{\mathrm{SG}}$, and F1$_{\mathrm{Rel}}$ measure structural agreement between a question-conditioned executed trajectory and one or more annotated reference evidence paths.
The two approaches are therefore complementary rather than interchangeable.

\subsection{Structural and Traversal Diagnostics}
\label{app:structural_metrics}

\paragraph{Structural and traversal statistics.}
For graph-level topological statistics, we construct an undirected simple entity graph by projecting each KG triple $(h,r,t)$ to an undirected edge $\{h,t\}$.
Parallel edges induced by different relation types between the same entity pair are therefore collapsed.
\emph{Graph density} is computed on this projection and measures its overall sparsity: higher values indicate a more densely connected entity graph, while lower values indicate a sparser graph.
The \emph{clustering coefficient} is likewise computed on the undirected simple projection and measures local neighborhood cohesiveness: higher values indicate more tightly connected local regions with cycles and short alternative paths, whereas lower values indicate a more tree-like local structure.
The number of \emph{weakly connected components} (WCCs) measures global graph connectivity after ignoring edge direction: $\mathrm{WCC}{=}1$ corresponds to a connected KG, while larger values indicate increasing fragmentation into isolated subgraphs.
In contrast, \emph{average out-degree} is computed on the original directed, relation-labeled KG and denotes the mean number of outgoing triples per entity, providing a global measure of the directed branching structure exposed by the KG.

\emph{RW-Ans$_p$} denotes the mean exact $N$-step probability, across evaluation questions, that an unbiased random walk from the topic entity terminates at a valid answer entity.
At each step, the walk selects uniformly from the valid navigation actions under the corresponding evaluator action space.
\emph{RW-Ans$_{\mathrm{MC}}$} denotes the Monte Carlo estimate of the same quantity, using $B_{\mathrm{roll}}=100$ sampled walks per question.
We report the mean and standard deviation across three random seeds.
We also report the average number of available actions along the annotated reasoning path, or along a representative valid path consistent with the annotated relation chain when only relation-chain supervision is retained.
This statistic characterizes the local branching factor encountered along the reference-consistent trajectory.

A challenging navigation benchmark should therefore combine scale with structural ambiguity: many entities, relation types, and triples enlarge the global search space; fewer WCCs indicate less fragmentation; higher clustering and moderate density indicate greater potential for local cycles, shortcuts, and distractor routes; higher directed and query-local action branching increases step-wise decision ambiguity; and low random-walk answer reachability indicates that successful answering is unlikely under unguided traversal and instead requires stronger question-conditioned guidance.

\paragraph{Structural calibration references.}
To contextualize the scale of the path-fidelity metrics, we evaluate two non-learned references.
The same unbiased random-walk baseline used to estimate RW-Ans$_{\mathrm{MC}}$ in Tab.~\ref{tab:qa_stats} is also used as a structural calibration reference for the path-fidelity metrics in Tab.~\ref{tab:combined_metric}. 
It samples uniformly from the evaluator's valid action space under the same hop budget $N$ as the learned navigation agents.
Because the random walk does not assign model scores to its trajectories, there is no distinguished top-ranked trajectory.
Instead, for each question we sample $B_{\mathrm{roll}}=100$ independent walks, compute PED, RED, $\mathrm{F1_{SG}}$, and $\mathrm{F1_{Rel}}$ for each sampled trajectory using the same path-fidelity definitions as for the learned agents, and average each metric over the sampled walks.
These per-question Monte Carlo averages are then averaged across evaluation questions.
We repeat the procedure with three random seeds and report the mean and standard deviation of the resulting seed-level averages.
The \emph{answer-oracle shortest-path} reference is given the topic entity, the navigation graph, and the set of valid answer entities, but not the annotated reasoning path. 
It deterministically selects a shortest graph path from the topic entity to any valid answer, with ties resolved by a fixed ordering of outgoing actions. 
The resulting path is evaluated using the same single- and multi-answer path-fidelity protocols as the learned navigation agents.
Neither reference produces ranked answer predictions, so MRR and Hits@1 are undefined.

\section{Additional Dataset Details}
\label{app:dataset_extra}

\subsection{\textsc{Kinship}}
\label{app:kinship_extra}

\smallskip
\textbf{Relation Types.}
\textsc{Kinship} contains 12 kinship relation types:
\emph{father}, \emph{mother}, \emph{son}, \emph{daughter}, \emph{aunt}, \emph{uncle}, \emph{niece}, \emph{nephew}, \emph{husband}, \emph{wife}, \emph{brother}, and \emph{sister}.
The \cite{kinship_55} release contains 104 $(\text{person},\text{relation})$ query instances, some of which have multiple valid answer labels.
When each valid answer is expanded into a separate $(\text{head},\text{relation},\text{tail})$ triple, these instances yield the 112 KG triples used in our graph representation.
Specifically, eight of the 104 query instances have two valid tail entities.

\smallskip
\textbf{Question-template variants.}
For each relation sequence $(r_1,\ldots,r_n)$ starting from topic entity $A$, we construct two chain renderings:
\begin{align*}
\textsc{Possessive}(A,r_{1:n}) 
&= \text{``$[A]$'s $[r_1]$'s $\cdots$'s $[r_n]$''},\\
\textsc{OfChain}(A,r_{1:n}) 
&= \text{``$[r_n]$ of the $\cdots$ of the $[r_1]$ of $[A]$''}.
\end{align*}
The possessive form follows the path order, while the ``of''-chain realizes the same relation sequence in reverse surface order.

\begin{table}[h]
\centering
\caption{Representative \textsc{Kinship} paraphrase-template families.}
\label{tab:kinship_template_examples}
\begin{tabular}{l|l}
\toprule
\textbf{Template family} & \textbf{Example realization} \\
\midrule
Possessive identity 
& ``Who is $[A]$'s $[r_1]$'s $\cdots$'s $[r_n]$?'' \\
Possessive name-seeking 
& ``What is the name of $[A]$'s $[r_1]$'s $\cdots$'s $[r_n]$?'' \\
Possessive conversational 
& ``Can you tell me who $[A]$'s $[r_1]$'s $\cdots$'s $[r_n]$ is?'' \\
Mixed final-relation form 
& ``Who is the $[r_n]$ of $[A]$'s $[r_1]$'s $\cdots$'s $[r_{n-1}]$?'' \\
Nested ``of''-chain 
& ``Who is the $[r_n]$ of the $\cdots$ of the $[r_1]$ of $[A]$?'' \\
Nested ``of'' conversational 
& ``Do you know who the $[r_n]$ of the $\cdots$ of the $[r_1]$ of $[A]$ is?'' \\
Family-member variant 
& ``Which family member is the $[r_n]$ of the $\cdots$ of the $[r_1]$ of $[A]$?'' \\
\bottomrule
\end{tabular}
\end{table}

For multi-hop paths, this procedure yields 23 unique paraphrase candidates. 
For one-hop paths, some possessive and ``of''-chain templates collapse to the same string after duplicate removal, yielding 19 unique candidates.
From these candidates, we randomly sample three question variants per path.

\smallskip
\textbf{Surface-order illustration.}
For the 3-hop chain
\[
\text{Penelope}
\xrightarrow{\text{husband}}
\text{Christopher}
\xrightarrow{\text{son}}
\text{Arthur}
\xrightarrow{\text{nephew}}
\text{Colin},
\]
the fixed ``of''-chain original question is:
\begin{quote}
``Who is the nephew of the son of the husband of Penelope?''
\end{quote}
whereas a possessive-chain variant is:
\begin{quote}
``Who is Penelope's husband's son's nephew?''
\end{quote}

\smallskip
\textbf{Sample Question--Answer Pairs}
\begin{quote}
\underline{1-hop question:} ``Who is the husband of Penelope?'' 
— \emph{Christopher}
\\
\underline{2-hop question:} ``Who is the son of the husband of Penelope?'' 
— \emph{Arthur}
\\
\underline{3-hop question:} ``Who is the nephew of the son of the husband of Penelope?'' 
— \emph{Colin}
\end{quote}

\smallskip
\textbf{Template-reliant paraphrasing.}
Our paraphrase construction deliberately relies on manually specified template families rather than free-form generation with LLMs or general-purpose paraphrasing models.
This choice reflects the requirements of path-based evaluation: question variants should alter the surface form without changing the entity identity, relation semantics, or ordered relation sequence.
In pilot experiments, automatic paraphrasers occasionally introduced unintended semantic drift, such as expanding a name like ``John'' into ``John the Baptist'', omitting relation-specific information, or collapsing entity and relation semantics into generic placeholders such as ``entity'' or ``relation''.
Such changes can make the intended reasoning path ambiguous or incorrect.
We therefore generate question variants only through controlled templates, and apply the same template-reliant construction principle to \textsc{MQuAKE-ST}.

\subsection{\textsc{MQuAKE-ST}}
\label{app:mquake_st_extra}

\smallskip
\textbf{Motivation and quality issues.}
\textsc{MQuAKE-CF} provides a useful starting point because it includes multi-hop questions, question variants, annotated reasoning paths, and Wikidata-grounded entity and relation identifiers.
Specifically, entities are represented by Wikidata item identifiers, or QIDs, while relations are represented by Wikidata property identifiers, or PIDs.
However, several properties make it unsuitable for direct graph-navigation evaluation.
First, using only the triples appearing in the annotated paths produces a highly fragmented KG ($\mathrm{WCC}{=}194$), substantially restricting the navigation space.
Second, \cite{zhong2025mquake} observe that some question texts omit key information from the underlying triple chain, causing the natural-language question to under-specify the intended reasoning path and rendering evaluation ill-posed.
They also identify text-level question duplicates; in our setting, we retain such cases only when the same entity label corresponds to distinct Wikidata identifiers, e.g., London (Q92561; city in Canada) and London (Q84; city in England).
Beyond text-level duplicates, we further observe path-level duplicates, where the same annotated reasoning path is associated with different question texts.

\smallskip
\textbf{Static KG reconstruction.}
To construct a fixed and challenging navigation space, we augment the KG using the entities and relations in the annotated paths as Wikidata seeds.
For each seed entity, we query Wikidata for triples in which the entity appears either as the head or tail, expanding the initial seed set from roughly $9$K triples to over $5$M candidate triples.
We then prune the expanded graph using a frequency threshold of $40$, retaining entities and relations that appear at least $40$ times in order to control graph size and reduce sparsity from infrequently used entities and relations.
Before pruning, redirected QIDs are canonicalized to a single identifier, with preference given to the original \textsc{MQuAKE-CF} QIDs when available, preventing the same entity from being split across multiple IDs.
The original seed entities and relations are exempt from pruning so that annotated reasoning paths are not removed solely by the frequency filter.
However, we do not forcibly insert the original seed triples into the final KG, since \textsc{MQuAKE-CF} and our reconstruction are based on different Wikidata snapshots: \textsc{MQuAKE-CF} was collected around mid-2022, whereas our queries were performed in early 2025.
For time-sensitive facts---such as the head of state, office holders, spouses, team memberships, or organizational affiliations---forcing older triples into a newer KG could introduce temporal inconsistencies.

\smallskip
\textbf{Question-template reconstruction.}
Since not all original \textsc{MQuAKE-CF} questions are directly usable for navigation, we reconstruct the QA layer using verified relation-chain templates.
We extract improved question templates from~\cite{zhong2025mquake} and align them with the relation-chain sequence of each path.
Using GPT-5.4~\cite{OpenAIGPT54} together with relation metadata, we further polish the templates so that each question remains consistent with the intended relation semantics, while adding additional variants where possible, especially for one-hop questions, which otherwise contain only a single variant.\footnote{All templates are manually verified after GPT-assisted editing.}
We discard the unusable original question--path pairs and sample replacement paths from the reconstructed KG whose relation-chains match the verified templates.
To increase coverage, we additionally sample paths for relation chains that are underrepresented in the reconstructed QA set.
For each retained or newly constructed path, we ensure that the path is contiguous, present in the KG, and unique at the identifier level; we then randomly sample three template questions for that path and select one as the canonical question.
For higher-hop paths, when a verified template matches a shorter subchain of the relation sequence, we also instantiate the corresponding lower-hop question.
Thus, uniqueness is enforced path-wise using Wikidata identifiers rather than surface text.

\smallskip
\textbf{Answer settings and released metadata.}
\textsc{MQuAKE-ST} provides two evaluation settings:
\textsc{MQuAKE-ST Single Answer} and \textsc{MQuAKE-ST Multi Answer}.
In \textsc{Single Answer}, each topic entity and relation chain is constrained to yield a single semantically valid answer.
In \textsc{Multi Answer}, all semantically valid answer entities reachable from the same topic entity under the corresponding relation-chain semantics are retained.
As in \textsc{MetaQA}, an agent in the multi-answer setting is marked correct if it reaches any valid answer entity.

We also release the verified relation-chain templates together with node and relation metadata.
Entities are represented by Wikidata item identifiers, or QIDs, while relations are represented by Wikidata property identifiers, or PIDs.
The node metadata records each entity's QID, title, description, aliases, Freebase MID when available, Wikipedia URL, and forwarding/redirect information.
The relation metadata records each relation's PID, title, description, and aliases.
Each released one-hop relation or multi-hop relation chain has at least three manually verified question templates.
During QA construction, we randomly sample three template questions for each path.
These resources support human-readable inspection of the KG and allow the KGQA community to generate additional multi-hop questions over the same static graph.


\smallskip
\textbf{Metadata examples.}
The released metadata keeps both machine-readable Wikidata identifiers and human-readable descriptions.
For relations, the Wikidata property identifier (PID) $\mathrm{P112}$ corresponds to the relation \emph{founded by}.
Its description specifies that the property denotes the founder or co-founder of an organization, religion, place, or entity, and its aliases include variants such as \emph{founder}, \emph{co-founder}, \emph{established by}, \emph{started by}, and \emph{created by}.
For entities, the Wikidata item identifier (QID) $\mathrm{Q552230}$ corresponds to the entity \emph{Miu Miu}, described as an Italian fashion house.
Its metadata includes the Freebase MID \texttt{/m/0h01k9} and the associated Wikipedia URL.
These fields allow each question and path to be inspected through readable labels while preserving identifier-level uniqueness for evaluation.

\smallskip
\textbf{Template examples.}
For the one-hop relation $\mathrm{P112}$, corresponding to \emph{founded by}, the template bank includes:
\begin{quote}
``Who founded $[X]$?''
\\
``Who is the founder of $[X]$?''
\\
``Who established $[X]$?''
\end{quote}
For the two-hop relation chain $\mathrm{P112}\!\rightarrow\!\mathrm{P69}$, corresponding to \emph{founded by} followed by \emph{educated at}, the template bank includes:
\begin{quote}
``Which educational institution educated the founder of $[X]$?''
\\
``What educational institution did the founder of $[X]$ attend?''
\\
``At which educational institution did the founder of $[X]$ receive their education?''
\end{quote}
For the three-hop relation chain $\mathrm{P112}\!\rightarrow\!\mathrm{P69}\!\rightarrow\!\mathrm{P159}$, the template bank includes variants such as:
\begin{quote}
``Where are the headquarters of the institution that educated the founder of $[X]$ located?''
\\
``Where is the educational institution attended by the founder of $[X]$ headquartered?''
\\
``In what place is the institution that educated the founder of $[X]$ headquartered?''
\end{quote}

\smallskip
\textbf{Single-answer illustration.}
In the \textsc{Single Answer} setting, a topic entity and relation chain are retained only when they yield a single semantically valid answer.
For example, the 3-hop chain
\[
\text{Miu Miu}
\xrightarrow{\text{founded by}}
\text{Miuccia Prada}
\xrightarrow{\text{educated at}}
\text{University of Milan}
\xrightarrow{\text{headquarters location}}
\text{Milan}
\]
induces the following prefix questions:
\begin{quote}
\underline{1-hop question:} ``Who established Miu Miu?''
--- \emph{Miuccia Prada}
\\
\underline{2-hop question:} ``At which educational institution did the founder of Miu Miu receive their education?''
--- \emph{University of Milan}
\\
\underline{3-hop question:} ``Where are the headquarters of the institution that educated the founder of Miu Miu located?''
--- \emph{Milan}
\end{quote}

\smallskip
\textbf{Multi-answer illustration.}
In \textsc{Multi Answer} setting, the same relation-chain semantics may produce multiple valid answer entities, all of which are retained.
For example:
\begin{quote}
\underline{1-hop question:} ``Who established Apple Inc.?''
--- \emph{Steve Jobs}; \emph{Steve Wozniak}
\\
\underline{1-hop question:} ``Who created Gwen Stacy?''
--- \emph{Stan Lee}; \emph{Steve Ditko}
\\
\underline{3-hop question:} ``Where are the headquarters of the organization that employed the founder of Apple Inc. located?''
--- \emph{Cupertino}; \emph{Palo Alto}; \emph{Emeryville}
\end{quote}
Thus, unlike \textsc{Single Answer}, the \textsc{Multi Answer} setting evaluates whether the predicted trajectory reaches any semantically valid terminal answer.




\section{Prior Path-Oriented KGQA Datasets}
\label{app:prior_dataset}

To contextualize the structural regime of the navigation-ready resources introduced in this work, we additionally characterize prior path-oriented KGQA benchmarks using the structural and traversal diagnostics defined in Appendix~\ref{app:structural_metrics}.
We consider \textsc{PathQuestion} (PQ), \textsc{PathQuestion-Large} (PQL)~\cite{zhou2018interpretable}, and the path-query subset of \textsc{WorldCup2014} (WC2014)~\cite{zhang2016gaussian}.
These resources provide natural-language questions associated with explicit reasoning paths and are therefore closely related to the question-conditioned navigation setting studied in this work.
Their released benchmark configurations nevertheless use compact, task-specific navigation KGs: PQ/PQL are distributed with hop-specific filtered Freebase KGs, whereas WC2014 operates over a small purpose-built football-domain KG.
Prior work has similarly noted that PQ/PQL use comparatively small KGs, contain repetitive relations with limited variety, and employ template-generated questions that can exhibit learnable patterns~\cite{chen2019uhop}.

\paragraph{Dataset source and counting convention.}
For reproducibility, we obtain PQ, PQL, and WC2014 from the public repository accompanying IRN~\cite{zhou2018interpretable},
\footnote{\url{https://github.com/zmtkeke/IRN}}
which provides the released question files and their corresponding task-specific KGs.
For PQ/PQL, the same artifacts are also distributed through an author-affiliated CCF-TCCI/Tsinghua release,
\footnote{\url{https://www.biendata.xyz/ccf_tcci2018/datasets/tcci_tag/16}}
providing a corroborating distribution of the files analyzed here.
For WC2014, the benchmark predates IRN, but the IRN repository is the earliest currently accessible public distribution of the derived KGQA artifacts that we were able to verify.
We therefore explicitly anchor all preprocessing and statistics below to the IRN release.

All statistics are computed directly from these files rather than transcribed from counts reported in prior work.
We count only entities and relation labels occurring in the materialized navigation KG and exclude model-specific vocabulary symbols such as \texttt{<unk>} and \texttt{<end>}.
For WC2014, the released \texttt{WC2014.txt} graph already materializes inverse predicates as distinct relation labels.
We retain and count these relations as part of the navigation graph and do not generate additional inverse edges.

\paragraph{Mixed-hop graph construction.}
For PQ and PQL, the public release provides separate filtered KGs for the 2-hop and 3-hop subsets.
\cite{zhou2018interpretable}, who introduced PQ and PQL, report 2,215 entities and 14 relations for PQ, and 5,035 entities and 364 relations for PQL.
\cite{qiu2020stepwise} subsequently evaluate mixed-hop settings denoted PQ-M and PQL-M, retaining these aggregate entity/relation counts and reporting 4,049 and 9,758 triples, respectively.
These published statistics do not match the graphs obtained directly from the released hop-specific KG files.
To make the mixed-hop construction explicit and reproducible, we therefore define
\[
    \mathcal{T}_{\mathrm{PQ-M}}
    =
    \mathcal{T}_{\mathrm{PQ-2H}}
    \cup
    \mathcal{T}_{\mathrm{PQ-3H}},
\]
and analogously for PQL, treating the union as a set so that triples appearing in both hop-specific KGs are counted only once.
Under this convention, \textsc{PQ-M} and \textsc{PQL-M} yield the graph statistics reported in Tab.~\ref{tab:kg_stats_extra}.
The released PQ-2H and PQ-3H KGs contain 1,211 and 2,839 triple rows, respectively, whose deduplicated union contains 3,377 unique triples.
For PQL, the 4,247-triple PQL-2H graph is fully contained in the 5,597-triple PQL-3H graph, so their union contains 5,597 unique triples.

For WC2014, the release directly provides the mixed-hop question file \texttt{WC-P.txt} together with the shared navigation KG \texttt{WC2014.txt}.
The mixed question file contains the same multiset of 6,482 1-hop and 1,472 2-hop examples as the corresponding hop-specific releases, although their ordering differs.
The original WC2014 paper reports 8,003 path-query instances, whereas the public IRN release used here contains 7,954 path-question instances, matching the count reported in~\cite{zhou2018interpretable}.
We therefore use the released mixed-hop file directly and report release-derived counts throughout.
The separate conjunctive-query subset of WC2014 is outside the sequential single-path setting considered here.
The resulting statistics characterize the actual graph exposed to the navigation agent rather than reproducing any particular aggregate dataset-size convention reported in prior work.

\begin{table}
    \caption{Structural statistics of prior path-oriented KGQA benchmarks under the
released navigation graphs used in our analysis.}
    \label{tab:kg_stats_extra}
      \centering
      \begin{tabular}{l | c c c}
        \toprule
        \textbf{Dataset} & \textbf{\textsc{WC2014-M}} & \textbf{\textsc{PQ-M}} & \textbf{\textsc{PQL-M}} \\
        \midrule
        Entities $|\mathcal{E}|$ & 1{,}127 & 2{,}256 & 6{,}505 \\
        Relations $|\mathcal{R}|$ & 10 & 13 & 411 \\
        Triples  $|\mathcal{T}|$ & 6{,}482 & 3{,}377 & 5{,}597 \\
        Avg. Out-Degree & 5.7516 & 1.4969 & 0.8604 \\
        WCCs & 1 & 46 & 1{,}346 \\
        Graph Density & \num[round-mode=places, round-precision=2, scientific-notation=true]{6.258E-03} & \num[round-mode=places, round-precision=2, scientific-notation=true]{1.240E-03} & \num[round-mode=places, round-precision=2, scientific-notation=true]{2.634E-04} \\
        Clustering Coefficient  & \num[round-mode=places, round-precision=2, scientific-notation=true]{2.752E-01} & \num[round-mode=places, round-precision=2, scientific-notation=true]{6.325E-02} & \num[round-mode=places, round-precision=2, scientific-notation=true]{6.030E-04} \\
        \bottomrule
      \end{tabular}
\end{table}

\paragraph{Traversal protocol.}
The released PQ/PQL files provide, for each row, a designated answer, a set of valid answers, and one corresponding reference path.
We use the released valid-answer set directly for answer-level evaluation and retain the relation chain of the annotated path for reference-consistent traversal diagnostics.
Paraphrased or otherwise distinct natural-language realizations are never merged and remain separate QA instances.
Rows with identical question text but different topic entities or relation chains are likewise retained as distinct instances, since they correspond to different navigation semantics.

For PQ, all released rows are retained as separate evaluation instances.
Each row therefore preserves its released natural-language realization and reference relation chain, while termination at any entity in its released valid-answer set counts as successfully reaching a valid answer.
Among the 7,106 released PQ rows, only one pair has identical question text, topic entity, relation chain, and hop count but different designated answers.
Since this answer-specific duplication is limited to a single pair, we retain the released row-level instance set rather than collapsing rows, preserving the instance count and weighting of the public release.

For PQL, repeated answer-specific rows are more common.
To avoid weighting the same question, topic entity, and relation-chain combination multiple times solely because different designated answers are stored as separate rows, we merge only rows that have identical question text, topic entity, relation chain, and hop count, after verifying that the grouped rows specify the same released valid-answer set.
Rows with identical question text but different topic entities or relation chains remain separate instances, since they define different navigation problems.
In the released PQL data, eight question strings are associated with more than one relation chain; these are retained as separate instances because each relation chain specifies a different ordered sequence of KG transitions despite the identical surface wording.
Each grouped PQL instance is represented by its topic entity, released valid-answer set, and relation chain; when a structural diagnostic requires an entity-level path, a representative valid path consistent with that relation chain is reconstructed from the navigation graph.

For PQ and PQL, which are released without fixed train/dev/test partition assignments, we follow the 8:1:1 train/validation/test proportion used in the original IRN experiments~\cite{zhou2018interpretable} and subsequently adopted for the mixed-hop evaluation in~\cite{qiu2020stepwise}.
We construct deterministic 80/10/10 splits independently for each hop count using random seed 42 and subsequently combine the corresponding 2-hop and 3-hop partitions.
For PQL, splitting is performed only after answer-specific rows are grouped, so rows corresponding to the same question, topic entity, and relation chain cannot be assigned to different partitions.
PQ and PQL also contain multiple natural-language realizations associated with the same topic entity and relation chain, which are retained as separate QA instances following the released benchmark format.
This differs from our \textsc{Kinship} and \textsc{MQuAKE-ST} protocol, where paraphrased variants are reserved for a separate robustness evaluation rather than included as additional canonical test instances.

For WC2014, we first separate the released mixed-hop resource according to the hop length encoded by its annotated reasoning path and construct deterministic 80/10/10 splits independently for the 1-hop and 2-hop subsets, following the 8:1:1 partition convention used for WC2014 by \cite{zhou2018interpretable} and \cite{qiu2020stepwise}.
We then recombine the corresponding partitions.
WC2014 likewise provides a set of valid answers for each question and is evaluated as a multi-answer benchmark, where terminating at any valid answer counts as success.
As in Tab.~\ref{tab:qa_stats}, evaluation-instance counts are reported over the full resource, whereas traversal-difficulty statistics are computed only over the resulting test split.
For mixed-hop evaluation, the navigation horizon is fixed to the maximum annotated hop length of each benchmark.
RW-Ans$_p$, RW-Ans$_{\mathrm{MC}}$, and the average number of available actions are computed using the same evaluator action space and definitions as for \textsc{Kinship} and \textsc{MQuAKE-ST} in Appendix~\ref{app:structural_metrics}.
Given these differences in graph construction and evaluation protocol---including the treatment of repeated answer-specific rows and paraphrased questions in PQ/PQL and the two-hop maximum of WC2014---we use these resources as structural reference benchmarks rather than additional model-evaluation datasets.

\begin{table}
    \caption{Evaluation-instance counts and traversal difficulty of prior path-oriented KGQA benchmarks under the released navigation graphs used in our analysis.}
    \label{tab:qa_stats_extra}
    \centering
    \small
    \resizebox{\textwidth}{!}{
    \begin{tabular}{l c | r r r r | c c c}
        \toprule
        \multirow{2}{*}{\textbf{Dataset}}
        & \multirow{2}{*}{\textbf{Answer Type}}
        & \multicolumn{4}{c|}{\textbf{Evaluation Instances by Hop Length}}
        & \multicolumn{3}{c}{\textbf{Traversal Difficulty}} \\
        \cmidrule(lr){3-6}
        \cmidrule(lr){7-9}
        &
        & \textbf{1-Hop}
        & \textbf{2-Hop}
        & \textbf{3-Hop}
        & \textbf{Total}
        & $\textbf{RW-Ans}_p$
        & $\textbf{RW-Ans}_{\mathrm{MC}}$
        & \textbf{Avg. Actions} \\
        \midrule

        \textsc{WC2014-M} & Multi
        & 6{,}482 & 1{,}472 & --  & 7{,}954
        & \num[round-mode=places, round-precision=2, scientific-notation=true]{1.4742e-01}
        & \numstats[sample][round-mode=places, round-precision=2, scientific-notation=true]{0.147641,0.147340,0.147340}
        & \num[round-mode=places, round-precision=2]{3.4439e+01} \\

        \textsc{PQ-M} & Multi
        & -- & 1{,}908 & 5{,}198 & 7{,}106
        & \num[round-mode=places, round-precision=2, scientific-notation=true]{1.8009e-01}
        & \numstats[sample][round-mode=places, round-precision=2, scientific-notation=true]{0.180281,0.179747,0.181210}
        & \num[round-mode=places, round-precision=2]{3.8605e+00} \\

        \textsc{PQL-M} & Multi
        & -- & 1{,}168 & 954  & 2{,}122
        & \num[round-mode=places, round-precision=2, scientific-notation=true]{5.8266e-01}
        & \numstats[sample][round-mode=places, round-precision=2, scientific-notation=true]{0.587042,0.584413,0.584178}
        & \num[round-mode=places, round-precision=2]{3.5536e+00} \\

        \bottomrule
    \end{tabular}
    }
\end{table}

\paragraph{Structural comparison.}
Tables~\ref{tab:kg_stats_extra} and~\ref{tab:qa_stats_extra} show that the three prior benchmarks occupy compact but structurally distinct navigation regimes.
\textsc{PQ-M} and \textsc{PQL-M} both expose relatively small local action spaces, but differ markedly in their global graph structure.
In particular, \textsc{PQL-M} combines a large relation vocabulary with a sparse and highly fragmented graph, yet still exhibits high unguided answer reachability under its grouped multi-answer evaluation: across evaluation instances, an unbiased $N$-step walk terminates at a valid answer with an average probability of $58.3\%$.
\textsc{PQ-M} is less fragmented and exhibits lower, though still comparatively high, random-walk answer reachability.
Together, these results show that graph sparsity and relation-vocabulary size alone do not determine navigation difficulty; the reachable search space and answer multiplicity must also be considered.

\textsc{WC2014-M} presents a different structural profile.
Its graph is connected and more locally clustered than those of \textsc{PQ-M} and \textsc{PQL-M}, while exposing a larger action space along annotated paths.
Its reasoning depth is nevertheless limited to at most two hops, and its multi-answer setting still permits comparatively high unguided answer reachability.
Thus, greater local branching does not necessarily imply a more difficult navigation problem: connectivity, reasoning depth, and answer multiplicity jointly shape the effective search space.

The two datasets introduced in this work deliberately occupy complementary regimes.
\textsc{Kinship} provides a compact, interpretable sandbox with substantial local relational ambiguity despite its small scale.
Its unguided answer reachability is comparable in scale to that of \textsc{PQ-M} and \textsc{WC2014-M}, showing that larger graph size alone does not make valid endpoints less accessible.

At the opposite extreme, \textsc{MQuAKE-ST} operates over a substantially larger shared graph with much higher local branching and far lower unguided answer reachability.
Its multi-answer setting exposes 83.36 actions on average while yielding an RW-Ans$_p$ of only $6.08\times10^{-3}$, more than an order of magnitude below any of the prior benchmarks in Tab.~\ref{tab:qa_stats_extra}.
Taken together, \textsc{Kinship} and \textsc{MQuAKE-ST} therefore support complementary forms of analysis: the former enables controlled study in a small, interpretable graph, whereas the latter evaluates question-conditioned traversal in a large, highly branching shared KG where valid answers are rarely reached by unguided exploration.
The prior resources occupy different intermediate structural regimes, but all remain considerably more favorable to unguided answer reachability than \textsc{MQuAKE-ST}.
These statistics do not by themselves determine benchmark quality or semantic reasoning difficulty; rather, they characterize the structural search space in which path-navigation performance is measured.

\section{Additional Simulations}
\label{app:additional_simulations}

\begin{table}[b]
    \caption{
    %
    Per-hop Hits@1 performance of models trained on the mixed $n$-hop settings in Tab.~\ref{tab:combined_metric}.
    %
    Best valid results are bolded within each model group.
    }
    \label{tab:per_hop_metrics}
    \centering

    \resizebox{\textwidth}{!}{
    \begin{tabular}{l cc ccc ccc}
        
        \toprule
        & \multicolumn{2}{c}{\textbf{\textsc{Kinship}}} 
        & \multicolumn{6}{c}{\textbf{\textsc{MQuAKE-ST}}} \\
        \cmidrule(lr){2-3}
        \cmidrule(lr){4-9}
        
        \textbf{Answer Type}
        & \multicolumn{2}{c}{Single} & \multicolumn{3}{c}{Single} & \multicolumn{3}{c}{Multi} \\
        \midrule

        \textbf{Hop Size} 
        & 2-hop & 3-hop 
        & 2-hop & 3-hop & 4-hop 
        & 2-hop & 3-hop & 4-hop \\ 
        \midrule

        \footnotesize\textbf{KGQA Baselines} & & & & & & & & \\

        \textsc{EmbedKGQA}
        & \best{\numstats{0.973684,0.947368,0.973684}} & \best{\numstats{1.000000,0.904762,0.904762}}
        & \numstats{0.519824,0.500000,0.497797} & \numstats{0.607143,0.597619,0.619048} & \numstats{0.744318,0.761364,0.750000}
        & \numstats{0.802198,0.821978,0.808791} & \numstats{0.910256,0.914530,0.910256} & \best{\numstats{0.983425,0.972376,0.983425}} \\

        \textsc{TransferNet}
        & \numstats{0.7895,0.9474,0.9211} & \numstats{0.8730,0.8730,0.9524}
        & \numstats{0.3866,0.7137,0.6707} & \numstats{0.3381,0.4405,0.1310} & \numstats{0.0568,0.5057,0.3693}
        & \numstats{0.6637,0.4593,0.8989} & \numstats{0.6966,0.3547,0.6111} & \numstats{0.2928,0.4862, 0.8343} \\

        \textsc{ReaRev}
        & \numstats{1.0,0.7105,1.0} & \numstats{0.8413,0.7143,0.9365}
        & \best{\numstats{0.9934,0.9912,0.9923}} & \best{\numstats{0.9643,0.9690,0.9571}} & \best{\numstats{0.9830,0.9886,0.9943}}
        & \best{\numstats{0.9846,0.9736,0.9714}} & \best{\numstats{0.9359,0.9701,0.9359}} & \numstats{0.9503,0.9503,0.9392} \\

        \midrule

        \footnotesize\textbf{Adapted Path-based Models} & & & & & & & & \\

        \textsc{MINERVA}
        & \best{\numstats{0.9474,0.9737,0.9211}} & \best{\numstats{0.9365,0.9524,0.9206}}
        & \best{\numstats{0.9119,0.9086,0.8711}} & \best{\numstats{0.8762,0.8810,0.8500}} & \numstats{0.8239,0.7841,0.8352}
        & \best{\numstats{0.8923,0.8725,0.8681}} & \numstats{0.8846,0.7607,0.8333} & \best{\numstats{0.9227,0.9392,0.9227}} \\

        \textsc{MultiHopKG}
        & \numstats{0.8421, 1.0000, 0.9474} & \numstats{0.9048, 0.9048, 0.9206}
        & \numstats{0.7081,0.6982,0.7423} & \numstats{0.6881,0.6952,0.6976} & \best{\numstats{0.8011,0.8409,0.8182}}
        & \numstats{0.8330,0.8462,0.8286} & \best{\numstats{0.8462,0.8462, 0.8162}} & \numstats{0.9006,0.9116,0.9116} \\

        \textsc{SQUIRE}
        & \numstats{0.447368,0.578947,0.473684} & \numstats{0.349206,0.444444,0.269841}
        & \numstats{0.570485,0.579295,0.591410} & \numstats{0.611905,0.621429,0.604762} & \numstats{0.659091,0.687500,0.715909}
        & \numstats{0.589011,0.600000,0.639560} & \numstats{0.662393,0.623932,0.666667} & \numstats{0.723757,0.674033,0.740331} \\
        
        \bottomrule

      \end{tabular}
}
\end{table}

\paragraph{Per-hop evaluation.}
In addition to the main $n$-hop results on \textsc{Kinship} and \textsc{MQuAKE-ST}, we evaluate the same trained models on their corresponding per-hop test subsets in Tab.~\ref{tab:per_hop_metrics}.
These results separate performance by annotated reasoning depth and provide a finer-grained view of hop-specific behavior.
For the direct answer-prediction baselines, \textsc{EmbedKGQA} is strongest on \textsc{Kinship}, while \textsc{ReaRev} is generally strongest on \textsc{MQuAKE-ST}.
\textsc{TransferNet} remains competitive on \textsc{Kinship}, but is less stable on \textsc{MQuAKE-ST}, where its performance decreases more noticeably with hop depth, especially in the single-answer setting.
This is consistent with the role of these models as answer-ranking references rather than path-level comparators.

Among the adapted path-based agents, \textsc{MINERVA} provides the most consistently strong answer-reaching performance across reasoning depths, while \textsc{MultiHopKG} matches or exceeds it on selected hop-specific subsets.
On \textsc{Kinship}, both models remain comparatively stable across the 2-hop and 3-hop subsets, whereas \textsc{SQUIRE} achieves substantially lower performance.
On \textsc{MQuAKE-ST}, \textsc{MINERVA} remains robust across hop lengths in both answer settings, although the single-answer setting shows a gradual decrease as the annotated reasoning depth increases.
\textsc{MultiHopKG} exhibits greater hop-dependent variation, but becomes increasingly competitive at deeper reasoning depths and can surpass \textsc{MINERVA} on individual subsets.
\textsc{SQUIRE} shows a clearer tendency to improve with hop length on \textsc{MQuAKE-ST}, but remains below the stronger reinforcement-learning agents in answer-reaching performance.
Overall, the per-hop results suggest that \textsc{MINERVA} offers the most consistent answer-reaching behavior across reasoning depths, while \textsc{MultiHopKG} and \textsc{SQUIRE} exhibit stronger depth-dependent effects.
The path-fidelity results in Tab.~\ref{tab:combined_metric} are needed to determine whether these answer-reaching trajectories also align with the annotated evidence paths.
%

\begin{table*}[t]
    \centering
    \caption{
    Performance on \textsc{MetaQA} ($n$-hop) with hop budget $N=3$. 
    Mean $\pm$ std over three trials. 
    Higher is better for MRR and Hits@1. 
    Per-hop results are obtained from the same mixed-hop model.
    Best valid results are bolded within each model group.
    }
    \label{tab:metaqa_metrics}

\resizebox{\textwidth}{!}{
    \begin{tabular}{l c cccc}
        
        \toprule
        \multirow{2}{*}{\textbf{Model / Metric}}
         & \multirow{2}{*}{MRR~$\uparrow$}
         & \multicolumn{4}{c}{Hits@1~$\uparrow$ } \\
        \cline{3-6}

         &  & n-hop & 1-hop & 2-hop & 3-hop\\
        \midrule

        \footnotesize\textbf{KGQA Baselines} & & & & & \\
        
        \textsc{EmbedKGQA} 
        & \numstats{0.812755,0.815538,0.814083} & \numstats{0.797146,0.741829,0.739547} & \numstats{0.732945,0.739715,0.733459} & \numstats{0.777748,0.783136,0.783001} & \numstats{0.698511,0.700478,0.699213} \\

        \textsc{TransferNet} 
        & \numstats{0.9915,0.8397,0.9837} & \numstats{0.9871,0.8047,0.9797} & \numstats{0.9606,0.9629,0.9750} & \numstats{0.9924,0.8573,0.9887} & \numstats{1.0000,0.6397,0.9737} \\

        \textsc{ReaRev} 
        & \numstats{0.8849,0.8633, 0.8713} & \numstats{0.8752,0.8563,0.8612} & \numstats{0.5288,0.4545,0.4778} & \numstats{0.9990,0.9990,0.9987} & \numstats{0.9877,0.9875,0.9850}\\


        
        

        \midrule

        \footnotesize\textbf{Adapted Path-based Models} & & & & & \\
        
        \textsc{MINERVA}
        & \best{\numstats{0.9414,0.9421,0.9486}} & \best{\numstats{0.8967,0.8978,0.9094}} & \best{\numstats{0.9064,0.9089,0.9038}} & \best{\numstats{0.8876,0.8911,0.9074}} & \best{\numstats{0.8994,0.8970,0.9155}}\\
        
        \textsc{MultiHopKG} 
        & \numstats{0.7120, 0.7245, 0.7063} & \numstats{0.6278,0.6480,0.6176} & \numstats{0.3418,0.3439,0.3293} & \numstats{0.7064,0.7419,0.7064} & \numstats{0.7451,0.7620,0.7261}  \\

        \textsc{SQUIRE} 
        & \numstats{0.628750,0.636692,0.633253} & \numstats{0.527690,0.539099,0.536029} & \numstats{0.385644,0.395195,0.387454} & \numstats{0.534494,0.551506,0.547068} & \numstats{0.619588,0.626454,0.628065}\\

        \bottomrule
    
    \end{tabular}
    }
\end{table*} 

\paragraph{\textsc{MetaQA}.}
For reference, we also report performance on \textsc{MetaQA}, a common multi-hop KGQA benchmark, in Tab.~\ref{tab:metaqa_metrics}.
Like \textsc{Kinship} and \textsc{MQuAKE-ST}, \textsc{MetaQA} contains templated questions, but its KG is typically treated as undirected due to its limited relation types.
Accordingly, we evaluate the adapted navigation models on an undirected version of the graph by augmenting observed triples with inverse edges.
We use the same candidate-based evaluation protocol described in Sec.~\ref{sec:problem_formulation} and Appendix~\ref{app:rollout_ranking}, with $B_{\mathrm{roll}}=100$ candidate trajectories per question.
The KG embedding dimension is set to $d_{\mathrm{KG}}=100$.
For the direct answer-prediction baselines, \textsc{TransferNet} provides strong and consistent answer-ranking performance across hop lengths, while \textsc{ReaRev} performs particularly well on the 2-hop and 3-hop subsets but is less stable on 1-hop questions.
Among the adapted path-based agents, \textsc{MINERVA} achieves the strongest overall answer-reaching performance and remains notably stable across 1-, 2-, and 3-hop questions.
\textsc{MultiHopKG} shows a stronger dependence on reasoning depth: its performance is weakest on the 1-hop subset but improves markedly on the 2-hop and 3-hop subsets.
\textsc{SQUIRE} exhibits a similar improvement with increasing hop length, although its overall answer-reaching performance remains lower.
Together, these results indicate that \textsc{MINERVA} generalizes more uniformly across hop depths, whereas \textsc{MultiHopKG} and \textsc{SQUIRE} benefit more noticeably from the longer-hop structure of \textsc{MetaQA}.
%
Reference-based path-fidelity metrics are not reported because \textsc{MetaQA} does not provide per-question evidence paths.

\begin{table} 
    \centering
    \caption{
    Oracle answer-generation performance on \textsc{MQuAKE-ST} under the gold-evidence setting.
    For single-answer questions, the LLM is given the annotated gold path; for multi-answer questions, it is given all semantically valid evidence paths.
    Correctness is evaluated by exact match against the answer set.
    Full model descriptions are provided in Tab.~\ref{tab:model_size}.
    }
    \label{tab:oracle_metrics}

    \begin{tabular}{lcc}
        \toprule
        \textbf{Model / Answer Type} & \textsc{Single Answer} & \textsc{Multi Answer} \\
        \midrule
        \textbf{LLama3.1-Instruct} & 0.930 & 0.988 \\
        \textbf{Qwen2.5-Instruct}  & 0.909 & 0.963 \\
        \textbf{Gemma4}            & 0.997 & 0.985 \\
        \bottomrule
    \end{tabular}
\end{table}

\paragraph{Answerability under gold evidence.}
To assess whether the constructed reasoning paths contain sufficient evidence for answering the corresponding questions, we evaluate them under an oracle, or gold-evidence, setting in Tab.~\ref{tab:oracle_metrics}.
No navigation agent is used in this experiment.
Instead, the LLM is given the question together with the fully labeled ground-truth evidence connecting the topic entity to the answer.
For single-answer questions, this evidence consists of the annotated gold path.
For multi-answer questions, we provide the set of semantically valid paths, forming an expanded gold evidence subgraph whose terminal entities correspond to acceptable answers.
This represents a perfect-retrieval setting: the supporting evidence is assumed to have already been recovered from the KG, and the remaining task is only to generate a correct answer from that evidence.
Thus, the experiment provides both a practical upper-bound estimate under the chosen downstream generator for a navigation-based QA pipeline and serves as a sanity check that the annotated paths, labels, and answer format are internally consistent.
High performance in this setting supports the answerability of the constructed question--evidence pairs.

The results in Tab.~\ref{tab:oracle_metrics} show that all evaluated LLMs achieve high Exact Match (EM) performance when given the ground-truth evidence.
On the single-answer split, EM ranges from $0.909$ to $0.997$, while on the multi-answer split it ranges from $0.963$ to $0.988$.
For multi-answer questions, correctness is evaluated by exact match against at least one acceptable answer entity.
This suggests that, under a perfect-retrieval assumption, the questions are largely answerable from the provided evidence context and that the label format is interpretable by the answer generator.
The strong performance in the multi-answer setting is particularly important because the LLM must identify at least one valid answer from an expanded evidence subgraph containing multiple semantically valid paths.
Remaining errors may therefore reflect answer-generation or formatting limitations, rather than an absence of supporting evidence in the provided paths.

This perspective also suggests a practical role for navigation models within a grounded KGQA pipeline.
Rather than returning only an answer score, a navigation model produces a ranked set of candidate answer entities together with explicit KG paths supporting each candidate.
The oracle-evidence experiment above shows that, when the relevant evidence is available, downstream language models can recover the correct answer with high accuracy.
This shifts an important part of the system-level challenge to evidence retrieval: the navigator should place a valid answer, together with a useful supporting path, as high as possible in the candidate ranking.
Under this interpretation, Hits@1 measures whether the first answer--evidence hypothesis presented downstream is correct, while MRR measures how quickly a valid hypothesis appears as additional unique candidate answers are considered.
A stronger ranking could reduce the number of candidate answer--path pairs that must be passed to a downstream generator, verifier, or reranker, lowering the evidence-processing budget while preserving an explicit reasoning trace for each candidate.
The navigation metrics thus have a direct operational interpretation: they measure not only whether the graph search can recover an answer, but also how efficiently it can surface grounded answer hypotheses for subsequent reasoning or verification.

\begin{table}
    \caption{Model sizes. For KGQA models, parameter counts are reported after instantiating each model on \textsc{MQuAKE-ST}. Language encoders and LLMs are kept frozen throughout our experiments. For LLM baselines, parameter counts and weight formats follow the Ollama checkpoints used for inference.
    }
      \label{tab:model_size}
      \centering
      \begin{tabular}{l | c c c}
        \toprule
        \textbf{Model} & \textbf{\# Params} & \textbf{Trainable Params} & \textbf{Precision} \\
        \midrule
        
        \footnotesize\textbf{KGQA Baselines} & & & \\
        \textsc{EmbedKGQA} &  7.54M & 3.69M & FP32 \\
        \textsc{TransferNet} & 6.14M & 2.59M & FP32 \\
        \textsc{ReaRev} & 1.69M & 1.69M & FP32 \\
        
        \midrule
        
        \footnotesize\textbf{Adapted Path-based models} & & & \\
        \textsc{MINERVA} & 4.60M & 4.60M & FP32 \\
        \textsc{MultiHopKG} & 10.27M & 10.27M & FP32 \\
        \textsc{SQUIRE} & 4.44M & 4.44M & FP32 \\
        
        \midrule
        
        \footnotesize\textbf{Language Encoder} & & & \\
        \textsc{BERT-base-uncased} & 110M & 0 & FP32 \\

        \midrule
        
        \footnotesize\textbf{LLM} & & & \\
        \textsc{Gemma-4-E4B} & 8B (4.5B effective) & 0 & Q4\_K\_M \\
        \textsc{Qwen2.5-7B-Instruct} & 7.62B & 0 & Q4\_K\_M \\
        \textsc{Llama-3.1-8B-Instruct} & 8.03B & 0 & Q4\_K\_M \\
        \bottomrule
      \end{tabular}
\end{table}

\paragraph{Model size.}
We report model-size statistics in Tab.~\ref{tab:model_size}.
For KGQA and adapted path-based models, parameter counts are measured after instantiating each model on \textsc{MQuAKE-ST}, since the size of the entity and relation embedding tables depends on the KG.
Most adapted navigation models themselves remain lightweight: \textsc{MINERVA} and \textsc{SQUIRE} contain $4.60$M and $4.44$M trainable parameters, respectively. 
\textsc{MultiHopKG} is twice the size at $10.27$M trainable parameters.
These sizes are comparable to the direct KGQA baselines, such as \textsc{EmbedKGQA} with $3.69$M trainable parameters, \textsc{ReaRev} with $1.69$M trainable parameters, and \textsc{TransferNet}  with 2.6M parameters.
Thus, the adapted path-based models do not rely on substantially larger trainable parameter budgets than the non-path-explicit KGQA baselines.

The main additional component is the frozen language encoder used to represent the question.
In our experiments, this is \textsc{BERT-base-uncased}, which has $110$M parameters.
However, this encoder is used only as a fixed feature extractor: it is not fine-tuned and does not add trainable parameters to the navigation models.
Even when including this frozen encoder, the resulting systems are still substantially smaller than the LLMs used for oracle answer generation, which contain roughly $7$B--$8$B parameters in the checkpoints used for inference.
The LLMs are also kept frozen and are used only as answer generators under the gold-evidence setting.
This distinction is important because our main comparison concerns the learned KGQA modules and navigation policies, rather than fine-tuning large language models.

\paragraph{Compute.}
Model trainings runs were performed on available PCs rather than a single dedicated workstation.
Consequently, different machines were used throughout, depending on availability. 
Additionally, some of the compute times differ due to concurrent jobs on the same machine.
Computer A has a Intel-core i7-14700F CPU, NVIDIA GeForce RTX 4080 16GB GPU, and 64 GB of RAM.
Computer B has an Intel-core i7-14700 CPU, NVIDIA GeForce RTX 3060 GPU, and 64 GB of RAM.
Computer C has an AMD Ryzen 9 7950X CPU, NVIDIA GeForce RTX 5070Ti GPU, and 128GB of RAM. 
Training of \textsc{MINERVA} on \textsc{MetaQA} on computer B required 1 day 8 hours and 30 minutes. 
\textsc{MINERVA} on \textsc{Kinship} on computer A took 1 hour and 10 minutes. 
\textsc{MQuAKE-ST} training on computer A showed the largest difference in run times, ranging between 4 and 6 hours.

\section{Adapted KGQA Models}
\label{app:model_adaptations}

Here we summarize the model-specific changes needed to instantiate the navigation-based baselines under the question-conditioned formulation in Sec.~\ref{section:preliminaries}.
We focus only on implementation-level differences from the original KGC-oriented models.
The general distinction between KGC and KGQA graph navigation is defined in Sec.~\ref{section:preliminaries}, and the evaluation protocol is described in Sec.~\ref{sec:problem_formulation}.
The adapted implementations are available for
\textsc{MINERVA}\footnote{\url{https://github.com/HalcyonSolutions/MINERVA}},
\textsc{MultiHopKG}\footnote{\url{https://github.com/HalcyonSolutions/MultiHopKG-NLP}},
and
\textsc{SQUIRE}\footnote{\url{https://github.com/HalcyonSolutions/SQUIRE}}.

All three adapted models replace the original symbolic query interface with a natural-language question interface.
Specifically, the question is encoded using a frozen language encoder and mapped into the corresponding model space through a learned projection module.
The projected question representation is then used as the conditioning context, replacing the symbolic relation-query input used in the original KGC formulations.
We also use a fixed-environment KGQA protocol: instead of constructing KGC-style instances with query-specific edge masking, the models navigate over the observed KG provided for each dataset.
At evaluation, answer metrics are computed from the ranked unique terminal entities induced by the candidate trajectories, while path-fidelity metrics are computed on the highest-scoring trajectory.
Finally, graph directionality is treated as a dataset-level choice.
Whereas KGC settings often augment triples with inverse relations to account for incomplete KG evidence, our adapted models can operate either on the directed KG or on an inverse-edge-augmented undirected variant, allowing us to match the graph assumptions of each dataset while keeping the model architectures unchanged.

\subsection{MINERVA}

\textsc{MINERVA} was originally formulated for symbolic query answering of the form $(h,r,?)$~\cite{das2018go}.
In that setting, the policy is conditioned on a learned embedding of the query relation $r$, and the agent walks from the start entity $h$ to a candidate tail entity.
Our adaptation preserves \textsc{MINERVA}'s recurrent path-based policy and relation--entity action scoring, but replaces the relation-query embedding with the projected question representation.
Thus, the underlying sequential decision process remains MINERVA-like, while the policy is conditioned on $c=\tilde{\mathbf{z}}_q$ rather than on a symbolic relation $r$.
At evaluation, candidate trajectories can be generated either by policy sampling or beam search.
In our experiments, we use beam search and rank trajectories by their cumulative policy log-probability.

For comparison, prior KGQA evaluations of \textsc{MINERVA} used lightweight question interfaces.
\citet{qiu2020stepwise} represent the question by averaging word embeddings and report 55.2\% Hits@1 on \textsc{MetaQA} 3-hop, while \citet{cohen2020scalable} use the non-entity portion of the question as a \textsc{MINERVA}-style relation input and report 41.7\% Hits@1 on \textsc{MetaQA} 3-hop.
These values correspond to hop-specific 3-hop evaluations, rather than to the mixed-hop training protocol used in our main experiments.
As an implementation reference, our question-conditioned \textsc{MINERVA} reaches 94.3\% Hits@1 when trained and evaluated specifically on the 3-hop \textsc{MetaQA} split.
In Sec.~\ref{sec:exp}, however, we report the more general $n$-hop setting, where a single policy is trained on variable-hop questions and evaluated by hop subset.

\subsection{MultiHopKG}

\textsc{MultiHopKG} was originally introduced as a path-based KGC model for symbolic queries of the form $(h,r,?)$~\cite{lin2018multi}.
Like \textsc{MINERVA}, it learns a sequential policy that walks over the KG from the start entity to a candidate answer entity.
Its main distinction is that it augments the reinforcement--learning objective with embedding-based reward shaping and action dropout, using a pretrained KG embedding model to provide softer structural feedback and encourage more diverse exploration.
Our adaptation keeps the path-search policy and action-space machinery, but replaces the symbolic query relation with a projected question representation from the language encoder.
The policy continues to represent the current state and candidate actions using KG entity and relation embeddings, while the projected question representation provides the conditioning context for selecting among the available actions.
However, we do not use the pretrained KG embedding model's triple score as an active reward-shaping signal.
In the original formulation, this score evaluates the plausibility of a terminal prediction under the symbolic query $(h,r,?)$ and can provide a nonzero reward when the predicted entity is not the target.
In our question-conditioned setting, the symbolic relation $r$ is replaced by a projected representation of the full natural-language question, for which the pretrained triple-scoring function was not trained.
We therefore use sparse answer supervision: the terminal reward is one when the reached entity belongs to the valid answer set and zero otherwise.
Thus, the adapted model retains \textsc{MultiHopKG}'s question-conditioned path-search policy, KG-based action representations, and action dropout, but not its embedding-based reward shaping.
We first pretrain a \textsc{ConvE} model on the KG triples to obtain the entity and relation embeddings used to initialize the \textsc{MultiHopKG} policy, without using the \textsc{ConvE} scoring function for reward shaping.
Once \textsc{MultiHopKG} is trained under this formulation, candidate trajectories can be generated either by policy sampling or beam search.
In our experiments, we use beam search and rank trajectories by their cumulative policy log-probability.

\subsection{SQUIRE}

\textsc{SQUIRE} was originally formulated for symbolic KG reasoning, where the input is a structured query and the model generates an evidential path sequence over KG tokens~\cite{bai2022squire}.
Unlike \textsc{MINERVA} and \textsc{MultiHopKG}, which learn stochastic walking policies with reinforcement learning, \textsc{SQUIRE} casts path reasoning as supervised sequence generation.
Given a symbolic query, the model autoregressively predicts a sequence of relations and entities corresponding to a candidate reasoning path.
Our adaptation preserves this path-generation view, but conditions the decoder on the projected question representation instead of the symbolic query input.
Training uses the annotated reasoning paths as supervised target sequences.
At evaluation, candidate trajectories are decoded under KG constraints.
In our experiments, we use beam search and score decoded trajectories by their cumulative sequence log-probability.
Multiple decoded trajectories may terminate at the same entity; in this case, we retain the highest-scoring trajectory for that entity.
Answer metrics are then computed over the resulting ranked list of unique terminal entities, while path-fidelity metrics are computed separately on the highest-scoring trajectory and compare it against the annotated evidence path.

\section{KGQA Reproducibility}
\label{app:reproducibility}

\subsection{EmbedKGQA}
\begin{table}
    \caption{Performance on MetaQA\_FULL (Hits@1).}
    \label{tab:embedkgqa_reproduced}
      \centering
      \begin{tabular}{l | c c c}
        \toprule
        \textbf{Model} & \textbf{1-Hop} & \textbf{2-Hop} & \textbf{3-Hop} \\
        \midrule
        EmbedKGQA (Reported) & 0.975 & 0.988 & 0.948 \\
        EmbedKGQA (Reproduced) & 0.739 & 0.783 & 0.698 \\
        \bottomrule
      \end{tabular}
\end{table}

Reproducing the reported \textsc{EmbedKGQA} results is limited by the unavailability of the specific preprocessed \textsc{MetaQA\_FULL} files used in the authors' repository~\cite{saxena2020improving}.
Although the original \textsc{MetaQA} dataset is publicly available, the download link for the preprocessed version used by \textsc{EmbedKGQA} is inactive at the time of writing, and this issue has been reported in the official repository.\footnote{\url{https://github.com/malllabiisc/EmbedKGQA/issues/142}}
As a result, a strict reproduction of the reported numbers is not possible.
Instead, we base our experiments on our own preprocessing of the original \textsc{MetaQA} dataset~\cite{zhang2017variational}, where the 1-, 2-, and 3-hop questions are combined into a single mixed-hop setting.
The model architecture follows the original \textsc{EmbedKGQA} design, using \textsc{TuckER} embeddings for the KG and an LSTM-based question encoder.
As shown in Tab.~\ref{tab:embedkgqa_reproduced}, the reproduced results are substantially lower than the reported numbers.
We attribute this discrepancy to differences in the unavailable preprocessing pipeline, sensitivity to data construction and training details, and the additional difficulty of learning a single model over mixed-hop questions rather than training separate hop-specific models.

\subsection{ReaRev}

For the \textsc{MetaQA} experiments, we use the preprocessed data released by the authors, together with their reported hyperparameters.
The release also includes pretrained word embeddings.
As an initial sanity check, training and evaluating \textsc{ReaRev} on the 3-hop \textsc{MetaQA} split yields a Hits@1 of $98.8\%$.
However, we find that comparable performance can be obtained without the released pretrained word embeddings.
This is important for our mixed-hop setting because the embedding-training procedure is not included in the release, and the provided embeddings are hop-specific and therefore not directly compatible across hop lengths.
We therefore train and evaluate \textsc{ReaRev} in the mixed-hop setting, which is more challenging than the standard single-hop-size setup.
In this setting, Hits@1 decreases to approximately $86\%$.
As shown in Tab.~\ref{tab:metaqa_metrics}, this drop is not primarily due to degraded 3-hop performance, but rather to the difficulty of learning the 1-hop cases.

We emphasize that \textsc{ReaRev} is evaluated under a different input setting from our adapted path-based navigation models.
Rather than navigating over the full KG, \textsc{ReaRev} assumes a question-specific subgraph constructed around the topic entity.
For \textsc{MetaQA}, the released preprocessing constructs this subgraph by BFS expansion from the topic entity, yielding nearly 500 unique entities and roughly 1100 triples per question on average.
Although the gold answer is not guaranteed to appear in every subgraph, the resulting answer coverage is at least $99\%$.

To obtain a setup comparable to both the authors' preprocessing and our question-conditioned graph navigation task, we construct a question-specific subgraph for each \textsc{MQuAKE-ST} instance by performing BFS expansion from the topic entity up to depth 4, subject to a limit of 1000 triples.
The resulting subgraphs contain approximately 600 unique entities on average.
Since \textsc{MQuAKE-ST} provides annotated reasoning paths, we additionally ensure that at least one annotated path is included in the subgraph whenever available.
For instances with multiple annotated paths, we enforce the inclusion of only one path.
For \textsc{Kinship}, the KG consists of two small family subgraphs; therefore, we provide the family subgraph corresponding to the question instance.

Unless otherwise stated, we use the authors' hyperparameters reported for the 3-hop \textsc{MetaQA} split for all datasets.
The main exceptions are that we do not use the released pretrained word embeddings and that we adjust the KG embedding dimension to match each dataset-specific experimental setup.
For \textsc{Kinship}, we set $d_{\mathrm{KG}}=12$ and train for 100 epochs, matching the setup in Tab.~\ref{tab:combined_metric}.
For \textsc{MQuAKE-ST}, we set $d_{\mathrm{KG}}=100$ to match the setup in Tab.~\ref{tab:combined_metric}, rather than using the authors' reported \textsc{MetaQA} value of $d_{\mathrm{KG}}=50$.

\subsection{TransferNet}
\label{app:transfernet_reproducibility}

For \textsc{TransferNet}~\cite{shi2021transfernet}, we use the authors' released \textsc{MetaQA-KB} implementation, adapting it to run on \textsc{Kinship} and \textsc{MQuAKE-ST}.
We retain the original BiGRU question encoder and differentiable relation-graph propagation.
\textsc{TransferNet} does not use learned KG embeddings and therefore has no dataset-specific $d_{\mathrm{KG}}$
We use hidden dimensions of 1024, 128, and 512 for \textsc{MetaQA}, \textsc{Kinship}, and \textsc{MQuAKE-ST}, respectively.

The model also uses the annotated hop count as an auxiliary training target for its learned hop selector.
This information is not provided at inference, where the model predicts answer scores from only the question and topic entity.
For \textsc{MQuAKE-ST} Multi-Answer, we extend the answer representation to a multi-hot target containing all valid answer entities and evaluate the resulting entity ranking against this answer set.

Compared with the other models considered in this section, \textsc{TransferNet} was relatively straightforward to reproduce and adapt to our experimental setting, requiring only minor modifications for \textsc{Kinship} and \textsc{MQuAKE-ST}.
\footnote{The released repository additionally depends on the \textsc{MovieQA} \texttt{wiki.txt} corpus for its \textsc{MetaQA-Text} experiments.
The original download is no longer available, but it is still retrievable through the Internet Archive's Wayback Machine.
Note that this resource is not required for the \textsc{MetaQA-KB} experiments reported here.}

\subsection{SRN and IRN}
\label{app:srn_irn_reproducibility}

SRN~\cite{qiu2020stepwise} and IRN~\cite{zhou2018interpretable} are closely related to our setting because both expose step-wise reasoning behavior for multi-relation KGQA.
However, we were unable to reliably reproduce and adapt their training procedures to our datasets due to incomplete release artifacts and missing implementation details.

For SRN,\footnote{\url{https://github.com/DanSeb1295/multi-relation-QA-over-KG}} we were unable to identify an author-released implementation, and therefore examined the public implementation available at the linked repository.
This implementation depends on pretrained entity and relation embeddings that are required to run the model but are not included in the repository, reportedly due to their size.
Although the repository provides basic execution instructions, it does not provide sufficient information to reproduce the missing embedding pretraining step, including the pretraining objective, hyperparameters, expected embedding format, or procedure for constructing the corresponding files for new datasets.
These missing details make it unclear how to construct the required embedding files for our datasets and connect them to the main training pipeline.

For IRN,\footnote{\url{https://github.com/zmtkeke/IRN}} we examined the author-associated repository linked to the paper.
However, the release does not provide a dependency specification, Python version, or README-style documentation describing the software environment and training procedure.
This makes it difficult to recover the original experimental setup and reliably reproduce the reported results.

We therefore exclude SRN and IRN from the empirical comparison, while noting that their path-explicit designs make them highly relevant candidates for future reproducibility efforts.


\end{document}